\documentclass[10pt,twocolumn]{article}

\usepackage{times}
\usepackage{epsfig}
\usepackage{graphicx}
\usepackage{amsmath}
\usepackage{amssymb}

\usepackage[top=2.0cm,bottom=2.0cm,left=2.0cm,right=2.0cm]{geometry}

\usepackage[pagebackref=true,breaklinks=true,letterpaper=true,colorlinks,bookmarks=false]{hyperref}

\begin{document}

\title{Multilinear Wavelets: A Statistical Shape Space for Human Faces}

\author{Alan Brunton\\
Fraunhofer Institute for Computer Graphics Research IGD\\
Fraunhoferstra{\ss}e 5, 64283 Darmstadt\\
{\tt\small alan.brunton@igd.fraunhofer.de}
\and Timo Bolkart\hspace{1cm} Stefanie Wuhrer\\
Cluster of Excellence MMCI, Saarland University\\
Campus E1 7, 66123 Saarbr\"{u}cken, Germany\\
{\tt\small \{tbolkart|swuhrer\}@mmci.uni-saarland.de}
}

\maketitle

\definecolor{checkmecol}{rgb}{0.7,0.2,0.3}
\newcommand{\xxx}{{\color{checkmecol} \ttfamily \small \textbf{xxx}}}
\newcommand{\remark}[1] {{\color{checkmecol} \ttfamily \small \textbf{[#1]}}}

\newcommand{\qheading}[1]{\noindent\textbf{#1}}

\newcommand{\vecx}{\textbf{x}}
\newcommand{\vecy}{\textbf{y}}
\newcommand{\vecp}{\textbf{p}}
\newcommand{\vecn}{\textbf{n}}
\newcommand{\vect}{\textbf{t}}
\newcommand{\vecl}{\textbf{l}}
\newcommand{\vecs}{\textbf{s}}
\newcommand{\vecw}{\textbf{w}}
\newcommand{\WT}{D}
\newcommand{\invWT}{\WT^{-1}}
\newcommand{\simT}{R}
\newcommand{\facemesh}{\mathcal{X}}
\newcommand{\lmkSet}{\mathcal{L}}
\newcommand{\lmkSetmodel}{\lmkSet^{(m)}}
\newcommand{\ptcloud}{\mathcal{P}}
\newcommand{\lmkSetdata}{\lmkSet^{(d)}}
\newcommand{\lmkmodel}{\ell^{(m)}}
\newcommand{\lmkdata}{\ell^{(d)}}
\newcommand{\norm}[1]{\left\|#1\right\|_2}
\newcommand{\distfunc}[2]{\norm{#1-#2}}
\newcommand{\weightfunc}{\rho}
\newcommand{\rank}{\mbox{rank}}

\begin{abstract}
We present a statistical model for $3$D human faces in varying expression, which decomposes the surface of the face using a wavelet transform, and learns many localized, decorrelated multilinear models on the resulting coefficients. Using this model we are able to reconstruct faces from noisy and occluded $3$D face scans, and facial motion sequences. Accurate reconstruction of face shape is important for applications such as tele-presence and gaming. The localized and multi-scale nature of our model allows for recovery of fine-scale detail while retaining robustness to severe noise and occlusion, and is computationally efficient and scalable. We validate these properties experimentally on challenging data in the form of static scans and motion sequences. We show that in comparison to a global multilinear model, our model better preserves fine detail and is computationally faster, while in comparison to a localized PCA model, our model better handles variation in expression, is faster, and allows us to fix identity parameters for a given subject.
\end{abstract}


\section{Introduction}

Acquisition of 3D surface data is continually becoming more commonplace and affordable, through a variety of modalities ranging from laser scanners to structured light to binocular and multi-view stereo systems. However, these data are often incomplete and noisy, and robust regularization is needed. When we are interested in a particular class of objects, such as human faces, we can use prior knowledge about the shape to constrain the reconstruction. This alleviates not only the problems of noise and incomplete data, but also occlusion. Such priors can be learned by computing statistics on databases of registered 3D face shapes.

Accurate 3D face capture is important for many applications, from performance capture to tele-presence to gaming to recognition tasks to ergonomics, and considerable resources of data are available from which to learn a statistical prior on the shape of the human face (e.g.~\cite{BlanzVetter1999,BU-3DFE_2006,BU-4DFE_2008,Bosphorus_2008}).

In this paper, we propose a novel statistical model for the shape of human faces, and use it to fit to input 3D surfaces from different sources, exhibiting high variation in expression and identity, and severe levels of data corruption in the forms of noise, missing data and occlusions. We make the following specific technical contributions:
\begin{itemize}
	\item A novel statistical shape space based on a wavelet decomposition of 3D face geometry and multilinear analysis of the individual wavelet coefficients.
	\item Based on this model, we develop an efficient algorithm for learning a statistical shape model of the human face in varying expressions.
	\item We develop an efficient algorithm for fitting our model to static and dynamic point cloud data, that is robust with respect to highly corrupted scans.
	\item We publish our statistical model and code to fit it to point cloud data~\cite{bbsw:statmods:2013}.
\end{itemize}

Our model has the following advantages. First, it results in algorithms for training and fitting that are highly efficient and scalable. By using a wavelet transform, we decompose a high-dimensional global shape space into many localized, decorrelated low-dimensional shape spaces. This dimensionality is the dominant factor in the complexity of the numerical routines used in both training and fitting. Training on thousands of faces takes a few minutes, and fitting to an input scan takes a few seconds, both using a single-threaded implementation on a standard PC. 

Second, it allows to capture fine-scale details due to its local nature, as shown in Figure \ref{fig:NoisyDataEgs}, while retaining robustness against corruption of the input data. The wavelet transform decomposes highly correlated vertex coordinates into decorrelated coefficients, upon which multilinear models can be learned independently. Learning many low-dimensional statistical models, rather than a single high-dimensional model, as used in~\cite{BlanzVetter1999,Vlasic2005,BolkartWuhrer2013}, greatly reduces the risk of over-fitting to the training data; it avoids the curse of dimensionality. Thus, a much higher proportion of the variability in the training data can be retained in the model. During fitting, tight statistical bounds can be placed on the model parameters for robustness, yet the model can still fit closely to valid data points.

Third, it is readily generalizable and extendable. Our model requires \emph{no explicit segmentation} of the face into parts; the wavelet transform decomposes the surface hierarchically into overlapping patches, and the inverse transform recombines them. Unlike manually decomposed part-based models, eg.~\cite{kakadiaris_etal_2007_deformable_model,haar_veltkamp_2008,smet_vanGool_2010}, it requires no sophisticated optimization of blending weights and the decomposition is not class-specific. Further, it can be easily extended to include additional information such as texture.

\section{Related Work}

This work is concerned with learning 3D statistical shape models that can be used in surface fitting tasks. To learn a statistical shape model, a database of shapes with known correspondence information is required. Computing correspondences between a set of shapes is a challenging problem in general~\cite{Tam2013}. However, for models of human faces, correspondences can be computed in a fully automatic way using template deformation methods (e.g.~\cite{Mpiperis2008,Salazar2012}). 

The most related works to our work are part-based multilinear models that were recently proposed to model 3D human body shapes~\cite{Chen2013}. To define the part-based model, a segmentation of the training shapes into meaningful parts is required. This is done manually by segmenting the human models into body parts, such as limbs. Lecron et al.~\cite{Lecron2012} use a similar statistical model on human spines, that are manually segmented into its vertebrae. In contrast, our method computes a suitable hierarchical decomposition automatically, thereby eliminating the need to manually generate a meaningful segmentation.

Many statistical models have been used to analyze human faces. The first statistical model for the analysis of $3$D faces was proposed by Blanz and Vetter~\cite{BlanzVetter1999}. This model is called the morphable model, and uses Principal Component Analysis (PCA) to analyze shape and texture of registered faces, mainly in neutral expression. It is applied to reconstruct $3$D facial shapes from images~\cite{BlanzVetter1999} and $3$D face scans~\cite{Blanz2007,Patel2009}. Amberg et al.~\cite{amberg_etal_fg08} extend the morphable model to consider expressions, by combining it with a PCA model for expression offsets with respect to the neutral expression geometry. An alternative way to incorporate expression changes is to use use a multilinear model, which separates identity and expression variations. This model has been used to modify expressions in videos~\cite{Vlasic2005,Dale2011,Yang2012}, or to register and analyze $3$D motion sequences~\cite{BolkartWuhrer2013}. Multilinear models are mathematically equivalent to TensorFaces~\cite{VasilescuTerzopoulos2002} applied to $3$D data rather than images, and provide an effective way to capture both identity and expression variations, and thus in Section \ref{sec_eval} we compare to a global multilinear model and show that our model better captures local geometric detail.

Blanz and Vetter~\cite{BlanzVetter1999} manually segmented the face into four regions and learned a morphable model on each segment. The regions are fitted to the data independently and merged in a post-processing step. This part-based model was shown to lead to a higher data accuracy than the global morphable model. As part-based models are suitable to obtain good fitting results in localized regions, they have been used in multiple follow-up works, eg.~\cite{kakadiaris_etal_2007_deformable_model,haar_veltkamp_2008,smet_vanGool_2010}. While the model of Kakadiaris et al.~\cite{kakadiaris_etal_2007_deformable_model} shares some similarities with our model, they use a fixed annotated face model, and wavelet transforms to compare facial geometry images. In contrast, we learn multilinear models on subdivision wavelet coefficients.

All of the methods discussed so far model shape changes using global or part-based statistical models. In contrast, by applying a wavelet transform to the data first, statistical models can be constructed that capture shape variation in both a local and multi-scale way. Such wavelet-domain techniques have been used extensively for medical imaging~\cite{davatzikos_etal,nain_etal_MICCAI05,li_etal_CVPR07}, and Brunton et al.~\cite{Brunton2011} proposed a method to analyze local shape differences of 3D faces in neutral expression in a hierarchical way. This method decomposes each face hierarchically using a wavelet transform and learns a PCA model for each wavelet coefficient independently. This approach has been shown to capture more facial details than global statistical shape spaces. Hence, in Section \ref{sec_eval} we compare to a wavelet-domain approach and show that our model better captures expression variation.

We propose a method that combines this localized shape space with a multilinear model, thereby allowing to capture localized shape differences of databases of 3D faces of different subjects in different expressions.

\section{Multilinear Wavelet Model}
\label{sec_model}

Our statistical shape space for human faces consists of a multilinear model for each wavelet coefficient resulting from a spherical subdivision wavelet decomposition of a template face mesh. The wavelet transform takes a set of highly correlated vertex positions and produces a set of decorrelated wavelet coefficients. This decorrelation means that we can treat the coefficient separately and learn a distinct multilinear model for each coefficient. These multilinear models capture the variation of each wavelet coefficient over changes in identity and expression. In the following, we review the two components of our model.

\subsection{Second Generation Spherical Wavelets}
\label{sec_model_wavelets}
Spherical wavelets typically operate on subdivision surfaces~\cite{spherical_wavelets} following a standard subdivision hierarchy, giving a multi-scale decomposition of the surface. This allows coarse-scale shape properties to be represented by just a few coefficients, while localized fine-scale details are represented by additional coefficients. Second generation wavelets can be accelerated using the lifting scheme~\cite{sweldens_lifting_1996}, factoring the convolution of the basis functions into a hierarchy of local lifting operations, which are weighted averages of neighboring vertices. When combined with subsampling, the transform can be computed in time linear in the number of vertices.
The particular wavelet decomposition we use~\cite{bspline_subdiv_wavelets} follows Catmull-Clark subdivision, and has been used previously for localized statistical models in multiple application domains~\cite{li_etal_CVPR07,Brunton2011}. The wavelet transform is a linear operator, denoted $\WT$. For a $3$D face surface $\facemesh$, the wavelet coefficients are $\vecs = \WT\facemesh$.

\subsection{Multilinear Models}
\label{sec_model_multilinear}
To statistically analyze a population of shapes, which vary in multiple ways, such as identity and expression for faces, one can use a multilinear model. In general, one constructs a multilinear model by organizing the training data into an $N$-mode tensor, where the first mode is the vector representation of each training sample, and the remaining modes contain training samples varied in distinct ways. 

We organize our set of parametrized training shapes into a $3$-mode tensor $\mathcal{A} \in \mathbb{R}^{d_1 \times d_2 \times d_3}$, where $d_1$ is the dimension of each shape, and $d_2$ and $d_3$ are the number of training samples in each mode of variation; in our case, identity and expression. It would be straightforward to extend this model to allow for more modes, such as varying textures due to illumination changes, if the data were available. We use a higher-order Singular Value Decomposition (HOSVD)~\cite{DeLathauwer1997} to decompose $\mathcal{A}$ into
\begin{equation}
	\mathcal{A} = \mathcal{M} \times_2 \textbf{U}_2 \times_3 \textbf{U}_3,
	\label{eq:TensorDecomposition}
\end{equation}
where  $\mathcal{M} \in \mathbb{R}^{d_1 \times m_2 \times m_3}$ is a tensor called a multilinear model, and $\textbf{U}_2 \in \mathbb{R}^{d_2 \times m_2}$ and $\textbf{U}_3 \in \mathbb{R}^{d_3 \times m_3}$ are orthogonal matrices. The $i$-th mode product $\mathcal{M} \times_i \textbf{U}_i$ replaces each vector $\textbf{m} \in \mathbb{R}^{m_i}$ of $\mathcal{M}$ in the direction of $i$-th mode by $\textbf{U}_i \textbf{m} \in \mathbb{R}^{d_i}$. To compute the orthogonal matrix $U_2$, $\mathcal{A}$ is unfolded in the direction of $2$-nd mode to the matrix $\textbf{A}_{(2)} \in \mathbb{R}^{d_2 \times d_1d_3}$, where the columns of $\textbf{A}_{(2)}$ are the vectors of $\mathcal{A}$ in direction of $2$-nd mode. 

The decomposition in (\ref{eq:TensorDecomposition}) is exact, if $m_i = \rank(\textbf{U}_{(i)})$ for all $i$. If $m_i < \rank(\textbf{U}_{(i)})$ for at least one $i$, the decomposition approximates the data. This technique is called truncated HOSVD, and we use this to reduce the dimensionality of the training data.

The multilinear model represents a shape $\textbf{s} \in  \mathbb{R}^{d_1}$ by
\begin{equation}
\label{eqn_multilinear_gen}
	\textbf{s} \approx \overline{\textbf{f}} + \mathcal{M} \times_2 \textbf{w}^T_2 \times_3 \textbf{w}^T_3,
\end{equation}
where $\overline{\textbf{f}}$ is the mean of the training data (over all identities and expressions), and $\textbf{w}_2 \in \mathbb{R}^{m_2}$ and $\textbf{w}_3 \in \mathbb{R}^{m_3}$ are identity and expression coefficients. Varying only $\vecw_2$ changes identity while keeping the expression fixed, whereas varying only $\vecw_3$ changes the expression of a single identity.

\section{Training}
\label{sec_training}

\begin{figure*}[tp]
	\centering
	\begin{tabular}{c}
		\includegraphics[width=0.8\textwidth]{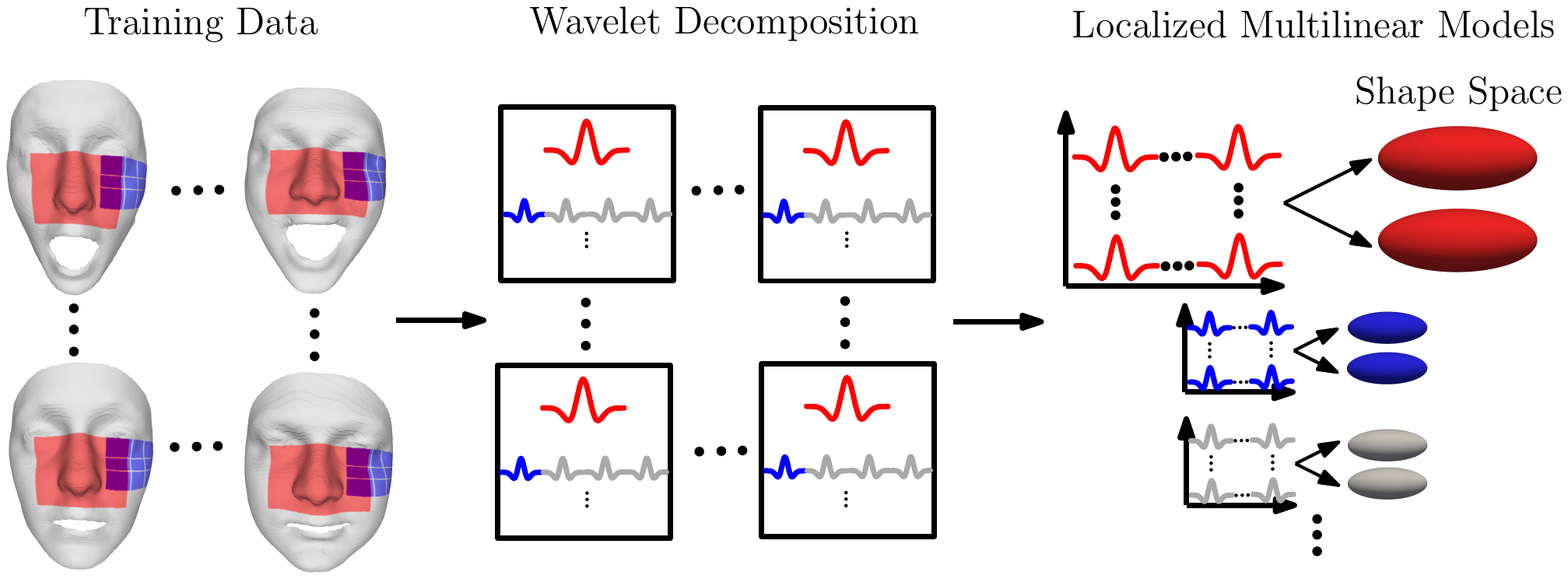}
	\end{tabular}
	\caption{Overview of the training. Left: Training data with highlighted impact of the basis function. Middle: Wavelet decomposition of each face of the training data. Right: Corresponding wavelet coefficients and learned multilinear model shape spaces.}			
	\label{fig:OverviewTraining}
\end{figure*}

In this section, we describe the process of learning the multilinear wavelet model from a database of registered 3D faces in a fixed number of expressions. Using the notation from Section \ref{sec_model_multilinear}, the database contains $d_2$ identities, each in $d_3$ expressions. We discuss in Section \ref{sec_eval} how to obtain such a registered database. The training process is depicted graphically in Figure \ref{fig:OverviewTraining}.

The first stage in our training pipeline is to apply a wavelet transform to every shape in our training database. The left-most part of Figure \ref{fig:OverviewTraining} shows the influence region of two wavelet coefficients on four face shapes (two identities in two expressions). To obtain a template with the proper subdivision connectivity, we use a registration-preserving stereographic resampling onto a regular grid~\cite{Brunton2011}, although any quad-remeshing technique could be used. Because the training shapes are registered, and have the same connectivity, we now have a database of registered wavelet coefficients (middle of Figure \ref{fig:OverviewTraining}). Note that this does \emph{not} require any manual segmentation, but is computed fully automatically. By considering the decorrelating properties of wavelet transforms, we can look at it another way: we now have a training set for each individual wavelet coefficient, which we can treat separately.

From these decorrelated training sets, covering variations in both identity and expression, we can learn a distinct multilinear model for each coefficient, resulting in many localized shape spaces as shown in the right part of Figure \ref{fig:OverviewTraining}. This allows a tremendous amount of flexibility in the model.

Training our model has the following complexity. Each wavelet transform has complexity $O(n)$, for $n$ vertices, and we perform $d_2 d_3$ of them. The complexity of the HOSVD is $O(d_1^2(d_2d_3^2+d_3d_2^2))$~\cite{DeLathauwer1997}, and we compute $n$ of them. Because every multilinear model is computed for only a single wavelet coefficient over the training set, $d_1=3$ so the complexity is $O(d_2 d_3^2 + d_3 d_2^2)$ per wavelet coefficient and $O(n (d_2d_3^2+d_3d_2^2))$ overall. 
Thus, our model allows highly efficient and scalable training, as detailed in Section \ref{sec_eval}.

Training many low-dimensional models has statistical benefits too. We retain a large amount of the variation present in the training data by truncating modes $2$ and $3$ at $m_2=3$ and $m_3=3$. We chose $m_2=m_3=3$ because $d_1=3$ is the smallest mode-dimension in our tensor.

Our model generates a $3$D face surface $\facemesh$ as follows. The vertex positions $\vecx\in\facemesh$ are generated from the wavelet coefficients via the inverse wavelet transform, denoted by $\invWT$. The wavelet coefficients are generated from their individual multilinear weights for identity and expression. Thus, following (\ref{eqn_multilinear_gen}), wavelet coefficients are generated by 
\begin{equation}
\label{eqn_recon_wavelet_multilinear}
\vecs_k=\overline{\vecs}_k + \mathcal{M}_k \times_2 \vecw^T_{k,2} \times_3 \vecw^T_{k,3} \\
\end{equation}
where $k$ is the index of the wavelet coefficient, and the surface is generated by
$\facemesh = \invWT\vecs$ 
where $\vecs=[\vecs_1\ \ldots\ \vecs_n]^T$.

\section{Fitting}
\label{sec_fitting}

\begin{figure}[tp]
	\centering
	\begin{tabular}{c}
		\includegraphics[width=0.4\textwidth]{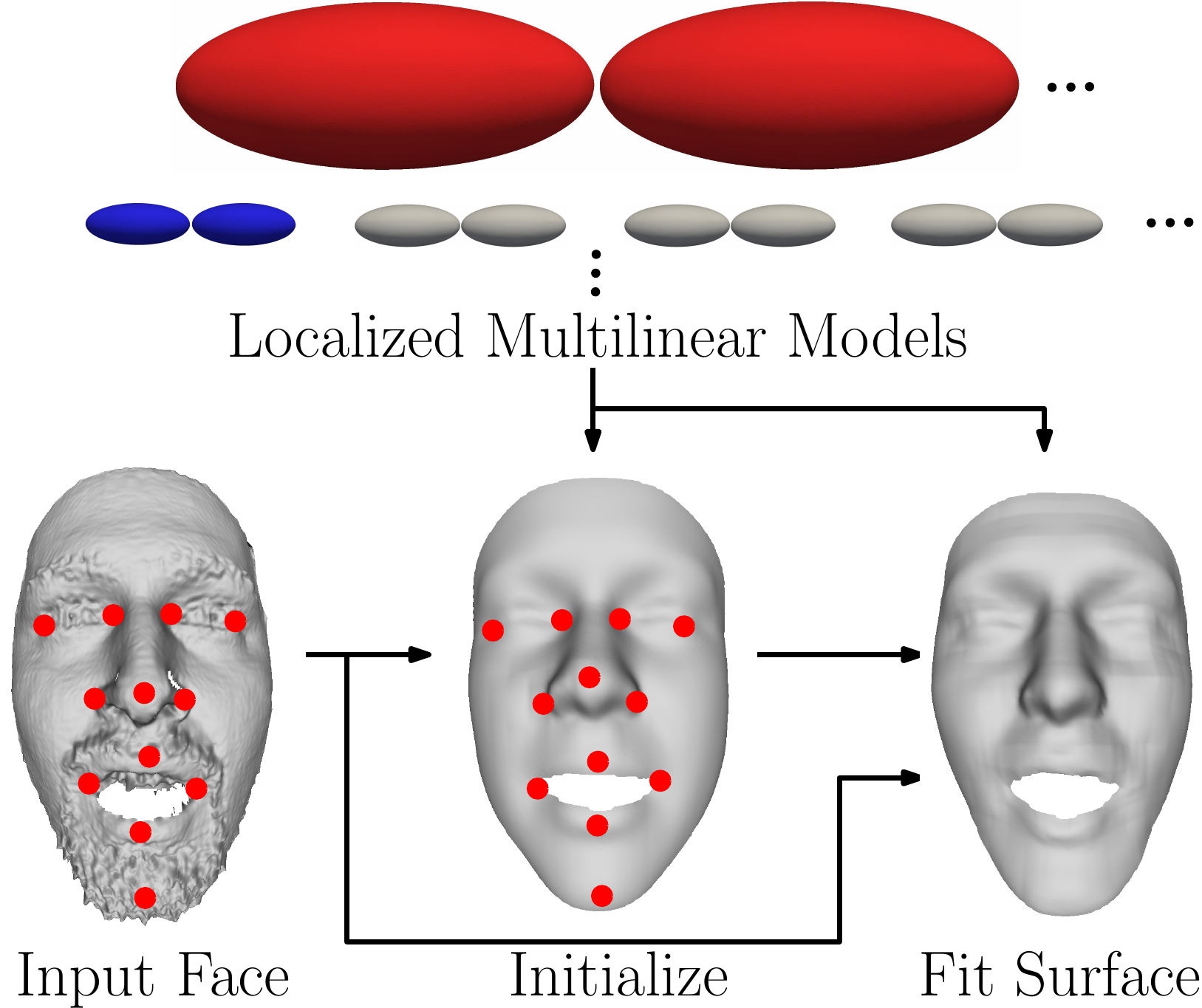}
	\end{tabular}
	\caption{Overview of the fitting. Top: Localized multilinear models. Bottom, left to right: input face scan, result after initialization, result of full surface fitting.}			
	\label{fig:OverviewFitting}
\end{figure}

In this section, we discuss the process of fitting our learned model to an input oriented point cloud or mesh $\ptcloud$, which may be corrupted by noise, missing data or occlusions. The process is depicted graphically in Figure \ref{fig:OverviewFitting}. We fit our model by minimizing a fitting energy that captures the distance between $\facemesh$ and $\ptcloud$, subject to the constraints learned in our training phase. We minimize the energy in a coarse-to-fine manner, starting with the multilinear weights of the coarse-scale wavelet coefficients, and refining the result by optimizing finer-scale multilinear weights.

\subsection{Fitting Energy}
\label{sec_fitting_energy}
We optimize our model parameters to minimize an energy measuring the distance between $\facemesh$ and $\ptcloud$. Our model parameters consist of the per-wavelet coefficient multilinear weights, $\vecw_{k,2}$, $\vecw_{k,3}$ for $k=1,\ldots,n$, and a similarity transform (rigid plus and uniform scaling) $\simT$ mapping the coordinate frame of $\facemesh$ to the coordinate frame of $\ptcloud$. 

Our fitting energy consists of four parts: a landmark term, a surface fitting term, a surface smoothing term, and a prior term. That is,
\begin{equation}
\label{eqn_fit_energy}
E_{\mbox{fit}} = E_{\lmkSet} + E_{\facemesh} + E_S + E_P
\end{equation}
where $E_{\lmkSet}$, $E_{\facemesh}$, $E_S$ and $E_P$ are the landmark energy, surface fitting energy, surface smoothing energy and prior energy, respectively. We now describe each of these energies in turn.

The landmark energy measures the Euclidean distance between corresponding landmark sets $\lmkSetmodel\subset\facemesh$ and $\lmkSetdata\subset\ptcloud$ located on the model surface and input data, respectively. These landmarks may be obtained in a variety of ways, including automatically~\cite{Creusot2013,Salazar2012}, and do not restrict our method. In Section \ref{sec_eval}, we demonstrate how our method performs using landmarks from multiple sources. The landmarks are in correspondence such that $|\lmkSetmodel|=|\lmkSetdata|$ and $\lmkmodel_i$ and $\lmkdata_i$ represent the equivalent points on $\facemesh$ and $\ptcloud$ respectively. With this, we define our landmark energy as,
\begin{equation}
\label{eqn_lmk_energy}
E_{\lmkSet} = \weightfunc_{\lmkSet} \frac{|\facemesh|}{|\lmkSetmodel|} \sum_{i=1}^{|\lmkSetmodel|} \distfunc{\simT\lmkmodel_i}{\lmkdata_i}^2
\end{equation}
where $\weightfunc_{\lmkSet}=1$ is a constant balancing the relative influence of landmarks against that of the rest of the surface.

The surface fitting energy measures the point-to-plane distance between vertices in $\facemesh$ and their nearest neighbors in $\ptcloud$. That is,
\begin{equation}
\label{eqn_surf_energy}
E_{\facemesh} = \sum_{\vecx\in\facemesh\backslash\lmkSetmodel} \weightfunc(\vecx) \distfunc{\simT\vecx}{\vecy(\vecx)}^2
\end{equation}
where $\vecy(\vecx)$ is the projection of $\simT\vecx$ into the tangent plane of $\vecp$, where $\vecp\in\ptcloud$ is the nearest neighbor of $\simT\vecx$. The distances are weighted by
\begin{equation}
\weightfunc(\vecx) = 
\begin{cases}
1 & \mbox{if } \distfunc{\simT\vecx}{\vecp} \leq \tau \\
0 & \mbox{otherwise}
\end{cases}
\end{equation}
where $\tau=1\mbox{cm}$ is a threshold on the distance to the nearest neighbor, providing robustness to missing data. We compute nearest neighbors using ANN~\cite{ANN}.

The prior energy restricts the shape to stay in the learned shape space, providing robustness to both noise and outliers. We avoid introducing undue bias to the mean shape via a hyper-box prior~\cite{BolkartWuhrer2013},
\begin{equation}
\label{eqn_prior_energy}
E_P = \sum_{k=1}^{n} \left( \sum_{j=1}^{m_2} f_{k,2,j}(w_{k,2,j}) + \sum_{j=1}^{m_3} f_{k,3,j}(w_{k,3,j}) \right)
\end{equation}
where 
\begin{equation}
f_{k,2,j}(w) = 
\begin{cases}
0				& \mbox{if } \bar{w}_{k,2,j}-\lambda \leq w \leq \bar{w}_{k,2,j}+\lambda \\
\infty	& \mbox{otherwise}
\end{cases}
\end{equation}
restricts each component of $\vecw_{k,2}$ to be within a constant amount $\lambda$ of the same component of the mode-mean $\bar{\vecw}_{k,2}$, and similarly for each component of $\vecw_{k,3}$.

The smoothing energy is the bi-Laplacian energy, which penalizes changes in curvature between neighboring vertices. It is needed due to the energy minimization algorithm, described in Section \ref{sec_fitting_min}, which optimizes each multilinear wavelet independently. Without a smoothing energy, this can result in visible patch boundaries in the fitted surface, as can be seen in Figure \ref{fig:smoothingDiff}.

Formally, we write
\begin{equation}
\label{eqn_smooth_energy}
E_S = \weightfunc_S \sum_{\vecx\in\facemesh} \norm{U^2(\vecx)}^2
\end{equation}
where $U^2(\vecx)$ is the double-umbrella discrete approximation of the bi-Laplacian operator~\cite{Kobbelt_multiresmodel_1998}, and $\weightfunc_S$ is a constant weight. 

The smoothing energy poses a trade-off: visually pleasing smooth surfaces versus fitting accuracy and speed. Leaving out $E_S$ allows the energy minimization to get closer to the data (as expected), and leads to faster fitting due to the energy being more localized. Hence, we retain the option of not evaluating this energy in case the scenario would favor close fitting and fast performance over visually smooth results. We use either $\weightfunc_S=100$ or $\weightfunc_S=0$ in all our experiments. Section \ref{sec_eval} discusses this trade-off in more concrete terms.

\subsection{Energy Minimization}
\label{sec_fitting_min}
We minimize (\ref{eqn_fit_energy}) in a two-step procedure. In the first step, we iteratively minimize $E_{\lmkSet} + E_P + E_S$ with respect to $\simT$ and the multilinear weights of each wavelet coefficient. This rigidly aligns the model and the data, and coarsely deforms the surface  to fit the landmarks, giving a good initialization for subsequent surface fitting. We solve for $\simT$ that minimizes $E_{\lmkSet}$, given the landmark positions $\lmkSetmodel$ and $\lmkSetdata$. This involves solving a small over-determined linear system. Then, we optimize $\vecw_{k,2}$ and $\vecw_{k,3}$ for $k=1,\ldots,n$ to minimize $E_{\lmkSet} + E_P$. Figure \ref{fig:OverviewFitting} (bottom, middle) shows the result of landmark fitting for a given input data.

In the second step, we fix $\simT$ and minimize (\ref{eqn_fit_energy}) with respect to only the multilinear weights. This deforms the surface so that it closely fits the input data $\ptcloud$. Figure \ref{fig:OverviewFitting} (bottom, right) shows the final fitting result.

The energies $E_{\lmkSet}$, $E_{\facemesh}$ and $E_S$ are nonlinear with respect to the multilinear weights, and we minimize them using the L-BFGS-B~\cite{LiuNocadal1989} quasi-Newton method. This bounded optimization allows the prior (\ref{eqn_prior_energy}) to be enforced simply as bounds on the multilinear weights. The hierarchical and decorrelating nature of the wavelet transform allows us to minimize the energies separately for each multilinear model in a coarse-to-fine manner. During initialization, we recompute $\simT$ and optimize the multilinear weights iteratively at each level of wavelet coefficients. During surface fitting, nearest neighbors are recomputed and the multilinear weights optimized iteratively at each level. During initialization, we allow greater variation in the model, $\lambda=1$, because we assume the landmarks are not located on occlusions. During surface fitting, we restict the shape space further, $\lambda=0.5$, unless the particular weight component is already outside this range from the initialization.

Fitting many low-dimensional local multilinear models is more efficient than fitting a single high-dimensional global multilinear model, because the dimensionality of the variables to be optimized is the dominant factor in the complexity of the quasi-Newton optimization, which achieves super-linear convergence by updating an estimate of the Hessian matrix in each iteration. For a problem size $d=m_2 + m_3$ the Hessian contains $\Omega(d^2)$ unique entries, which favors solving many small problems even if the total number of variables optimized is greater. This is confirmed experimentally in Section \ref{sec_eval}. Further, each multilinear model has compact support on $\facemesh$, which reduces the number of distances that must be computed in each evaluation of (\ref{eqn_surf_energy}) and its gradient.

\subsection{Tracking}
\label{sec_fitting_track}
As an application of our shape space, we show how a simple extension of our fitting algorithm can be used to track a facial motion sequence. To the first frame, we fit both identity and expression weights. Subsequently, we fix identity weights and only fit expression weights. This ensures that shape changes over the sequence are only due to expression, not identity. A more elaborate scheme, which averages the identity weights, would also be feasible.

To avoid jitter, we introduce a temporal smoothing term on the vertex positions. Approaches based on global multilinear models often place a temporal smoothing term on the expression weights themselves~\cite{Yang2012,BolkartWuhrer2013} since these are usually much lower dimension than the surface $\facemesh$. In our case, the combined dimensionality of all expression weights is equal to that of the vertex positions, so no efficiency is to be gained by operating on the weights rather than the vertex positions. Further, placing a restriction on the vertex positions fits easily into our energy minimization. We use a simple penalty on the movement of the vertices $\vecx\in\facemesh$ between frames. This is easily incorporated into our fitting algorithm by simply adding a Euclidean distance penalty to our energy function (\ref{eqn_fit_energy}) during surface fitting:
\begin{equation}
E_T = \sum_{\vecx_t\in\facemesh_t} \weightfunc_{T}\distfunc{\vecx_t}{\vecx_{t-1}}^2
\end{equation}
where $\weightfunc_{T}=1$ is a constant balancing allowing the surface to move versus reducing jitter.

\section{Evaluation}
\label{sec_eval}

\subsection{Experimental Setup}
\label{sec_eval_setup}
\qheading{Training Data:} For a training database, we use the BU3DFE database~\cite{BU-3DFE_2006} registered using an automatic template-fitting approach~\cite{Salazar2012} with ground truth landmarks. This database contains $100$ subjects in $25$ expressions levels each. We successfully registered $99$ subjects in all expressions and used this for training in our experiments.

\qheading{Test Data:} To test our fitting accuracy we use $200$ scans from the Bosphorus database~\cite{Bosphorus_2008} including variation in identity, expression and types of occlusions. We specifically do \emph{not} test on scans from the same database we use for training to avoid bias. Further, the Bosphorus scans typically have higher noise levels than those in BU3DFE, and contain occlusions. This database contains landmarks on each scan; we use the subset of those shown in Figure \ref{fig:OverviewFitting} present on a given surface (not blocked by an occlusion). In Section \ref{sec_eval_motion}, we show the performance of our method when tracking facial motion sequences from the BU4DFE database~\cite{BU-4DFE_2008} with landmarks automatically predicted using an approach based on local descriptors and a Markov network~\cite{Salazar2012}.

\qheading{Comparison:} We compare our fitting results to the localized PCA model~\cite{Brunton2011} and the global multilinear model~\cite{BolkartWuhrer2013}. All three models are trained with the same data, with the exception that because the local PCA model does not model expression variation, we train it separately for each expression and give it the correct expression during fitting. The other two are given landmarks for fitting.

\qheading{Performance:} We implemented our model, both training and fitting, in C++ using standard libraries. We ran all tests on a workstation running windows with an Intel Xeon E31245 at $3.3\mbox{GHz}$. Training our model on $2475$ face shapes each with $24987$ vertices takes $<5\mbox{min}$ using a single-threaded implementation. In practice we found our training algorithm to scale approximately linearly in the number of training shapes. Fitting takes $5.37\mbox{s}$ on average with $\weightfunc_S=0$, and $14.76\mbox{s}$ with $\weightfunc_S=100$, for a surface with approximately $35000$ vertices (Sections \ref{sec_eval_noisy} and \ref{sec_eval_occluded}). For the motion sequences with approximately $35000$ vertices per frame (Section \ref{sec_eval_motion}), fitting takes $4.35\mbox{s}$ per frame on average without smoothing and $11.14\mbox{s}$ with smoothing. The global multilinear model takes $\approx 2\ \mbox{min}$ for fitting to a static scan. A single-threaded implementation of the local PCA model takes $5\ \mbox{min}$ due to the sampling-based optimization, which avoids local minima.

\subsection{Reconstruction of Noisy Data}
\label{sec_eval_noisy}

\begin{figure}[t]
\centering
\parbox{0.45\textwidth}
{
	\includegraphics[width=0.09\textwidth]{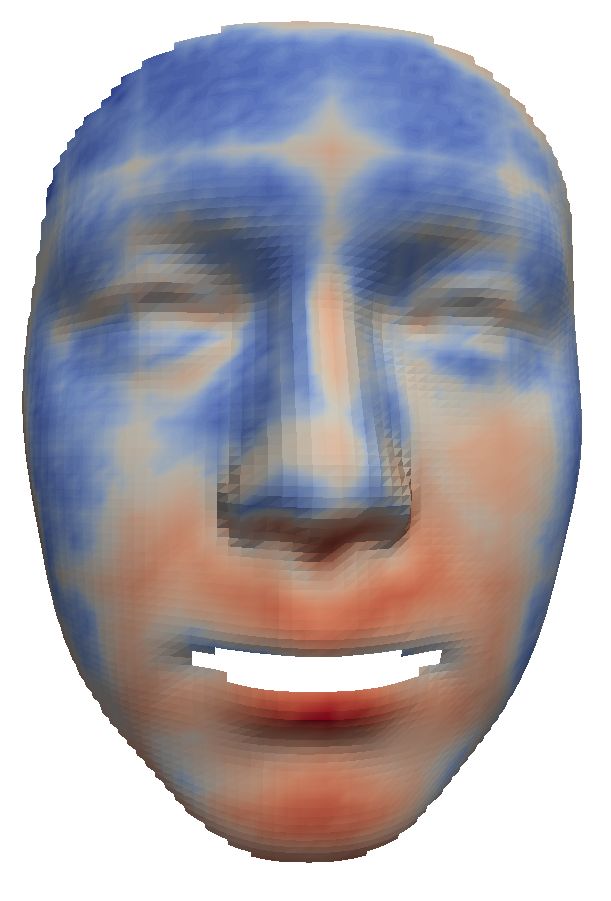}\hfill
	\includegraphics[width=0.09\textwidth]{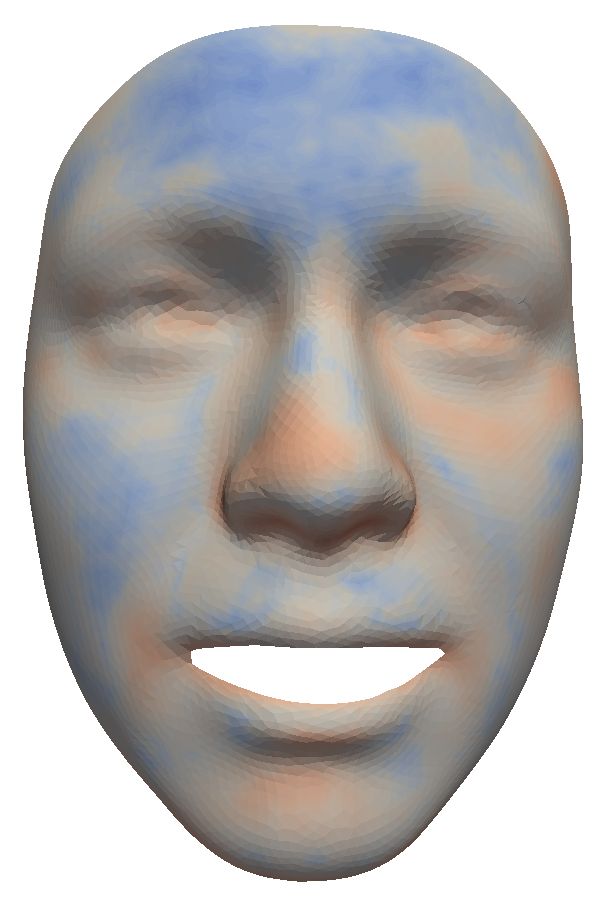}\hfill
	\includegraphics[width=0.09\textwidth]{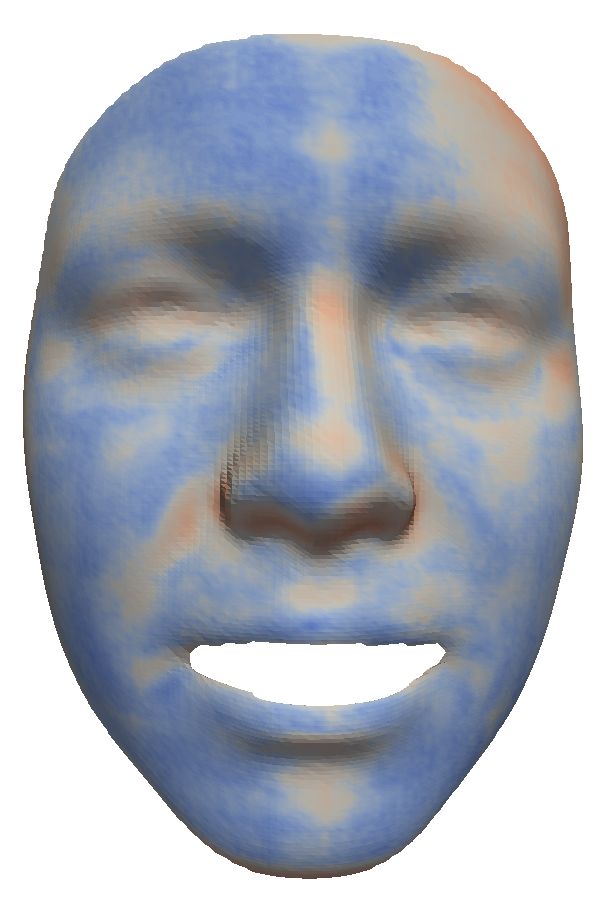}\hfill
	\includegraphics[width=0.09\textwidth]{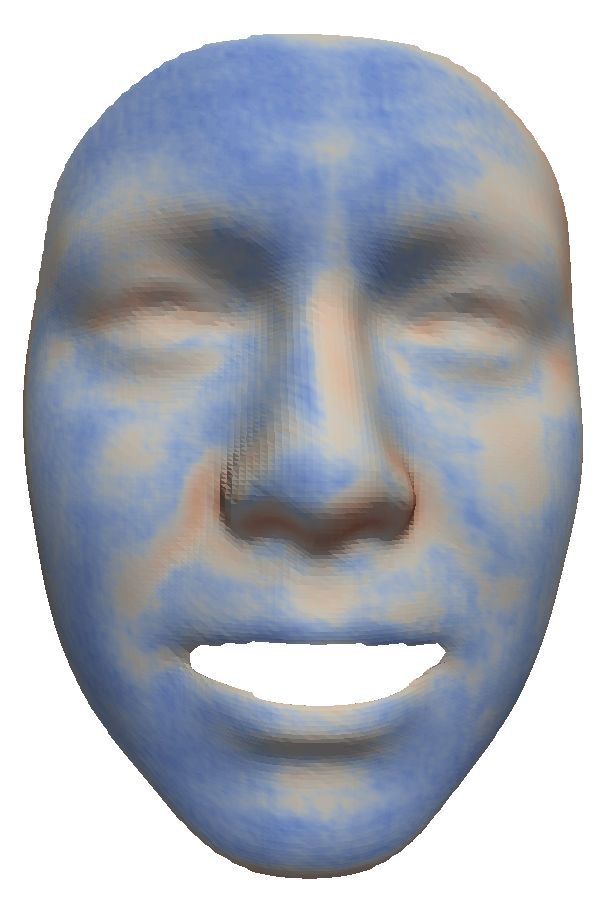}\hfill
	\includegraphics[width=0.05\textwidth]{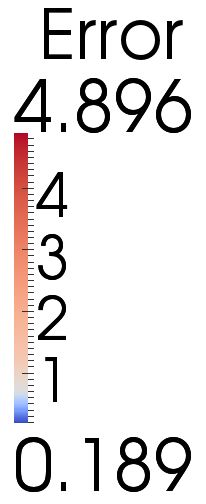}		
}
\\
\vspace{0.1cm}
\parbox{0.45\textwidth}
{
\raisebox{0.35cm}{
\hspace{0.5cm}
\includegraphics[width=0.11\textwidth]{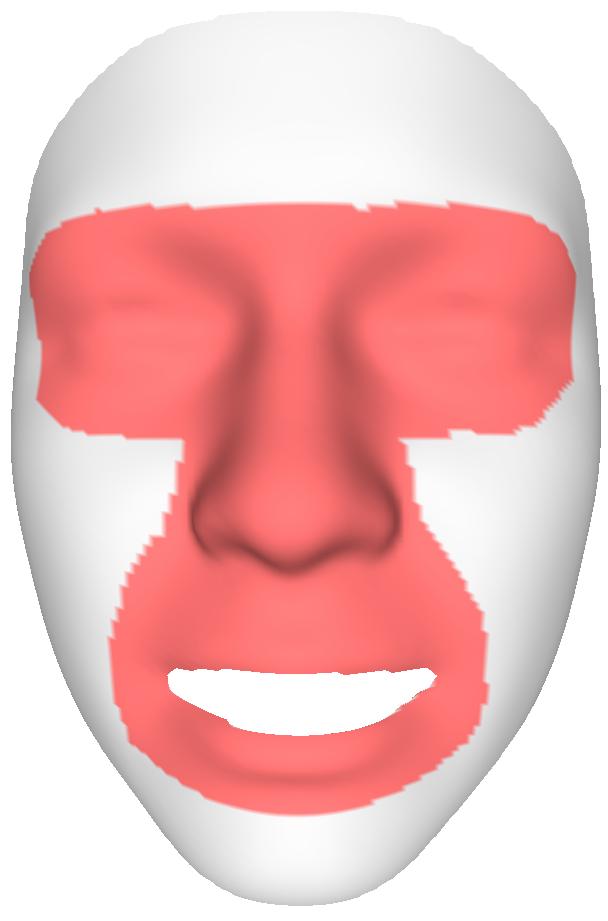}
}\hfill
\includegraphics[width=0.25\textwidth]{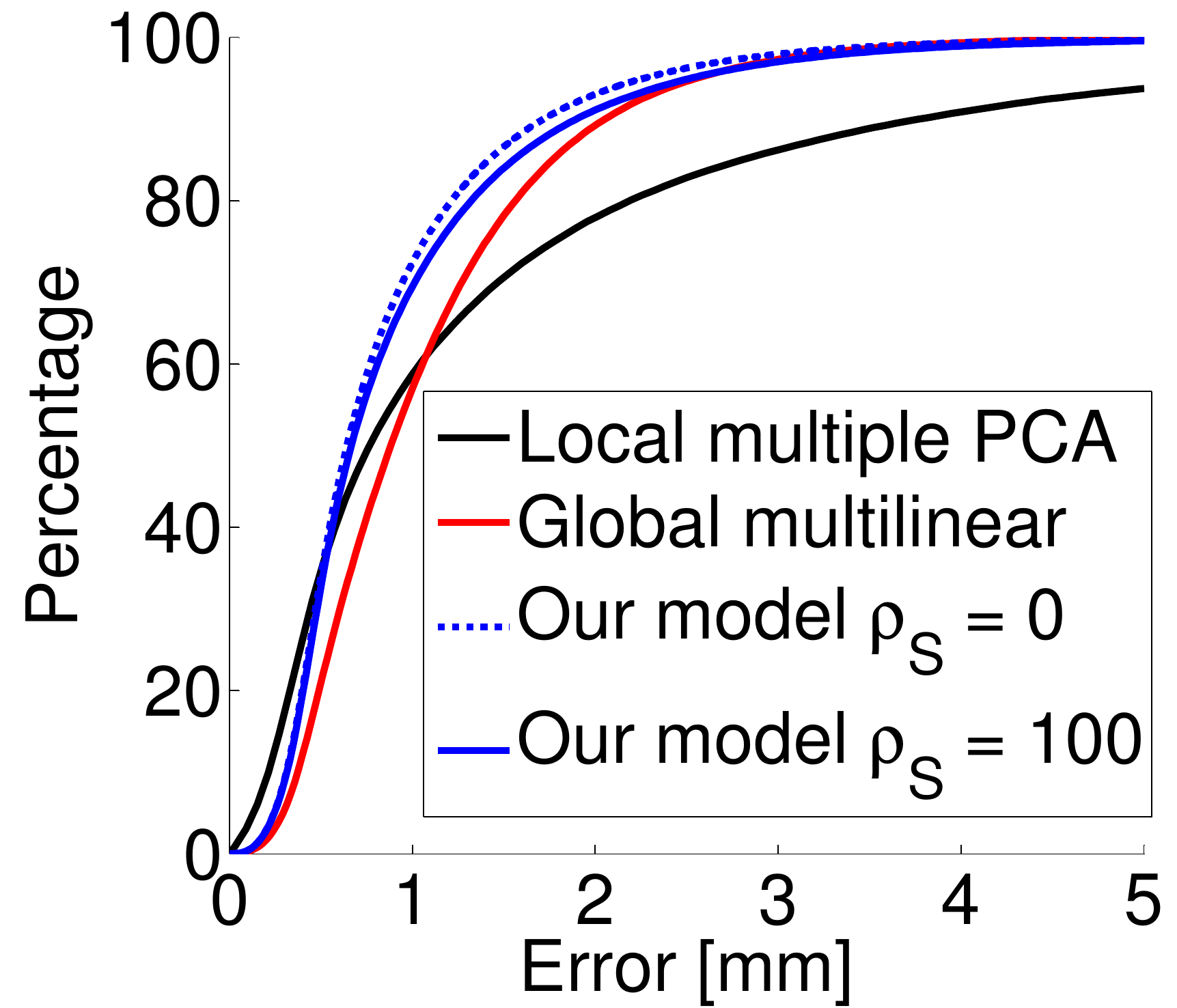}
}
\caption{Top block: Median reconstruction error for noisy data using multiple localized PCA models, a global multilinear model, our model ($\weightfunc_S=0$), and our model ($\weightfunc_S=100$). Bottom block: mask showing the characteristic detail regions of the face, and cumulative error plot for varying identity and expression. Errors in millimeters.}
\label{fig:NoisyDataQuant}
\end{figure}

\begin{figure*}[t]
\centering
\parbox{0.65\textwidth}
{
\includegraphics[width=0.125\textwidth]{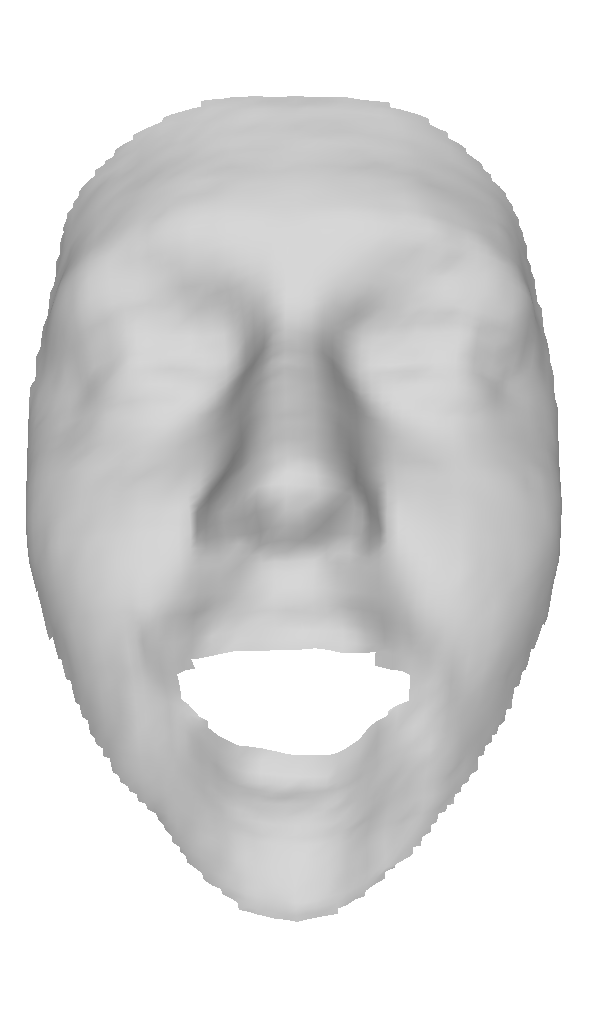}\hfill
\includegraphics[width=0.125\textwidth]{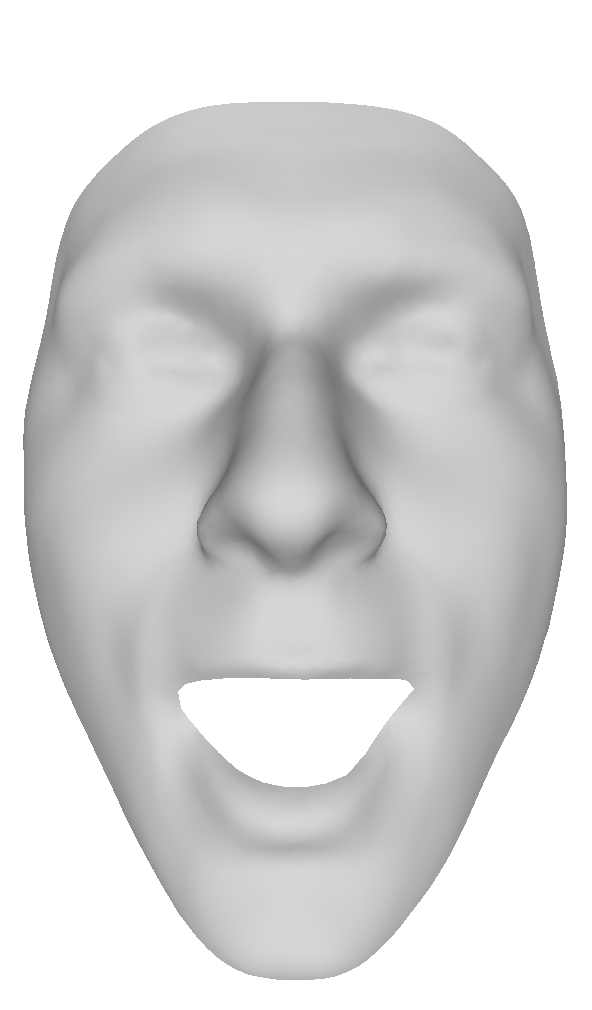}\hfill
\includegraphics[width=0.125\textwidth]{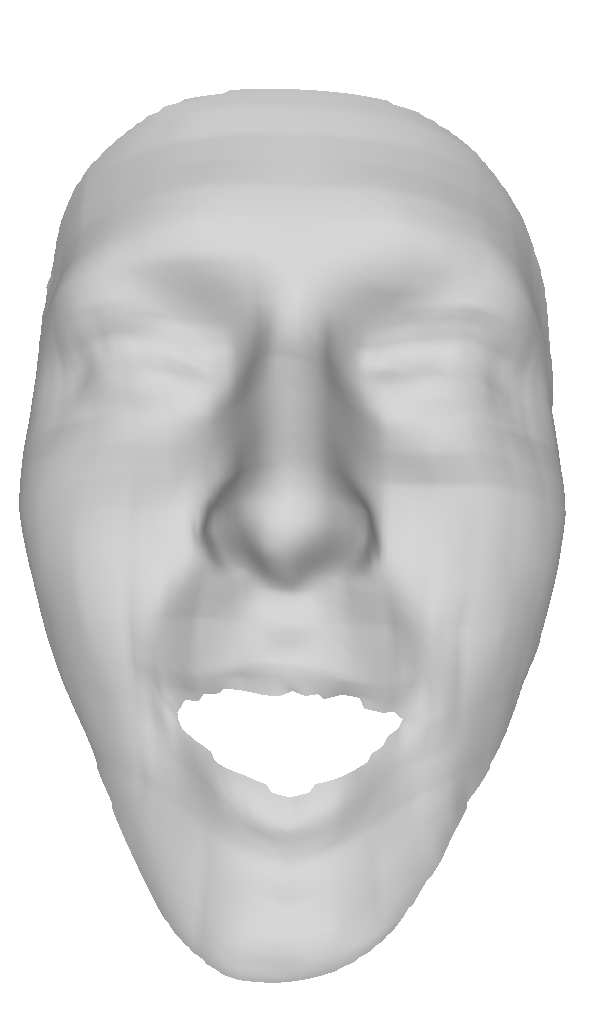}\hfill
\includegraphics[width=0.125\textwidth]{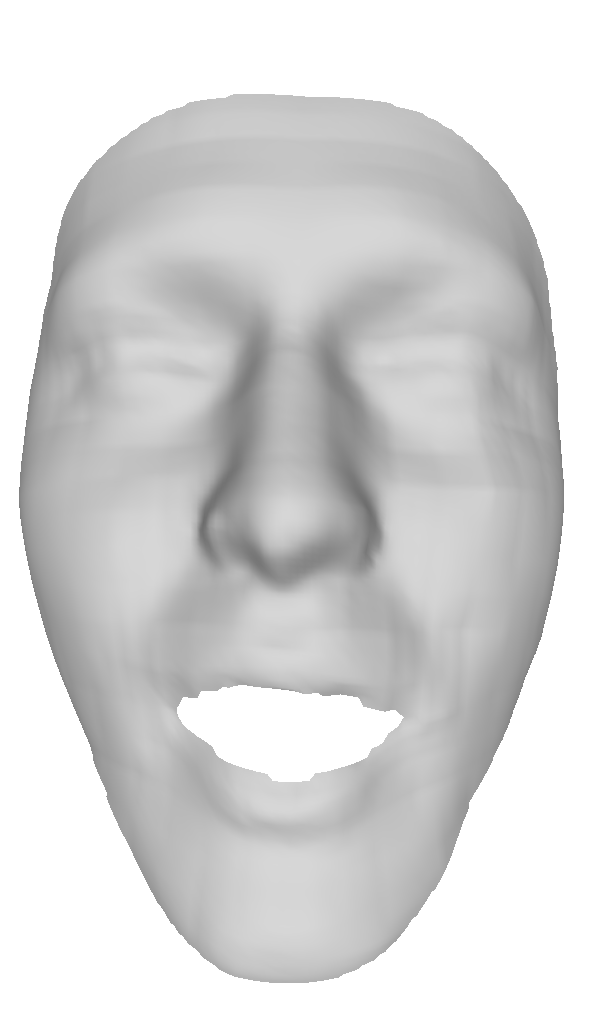}\hfill
\includegraphics[width=0.125\textwidth]{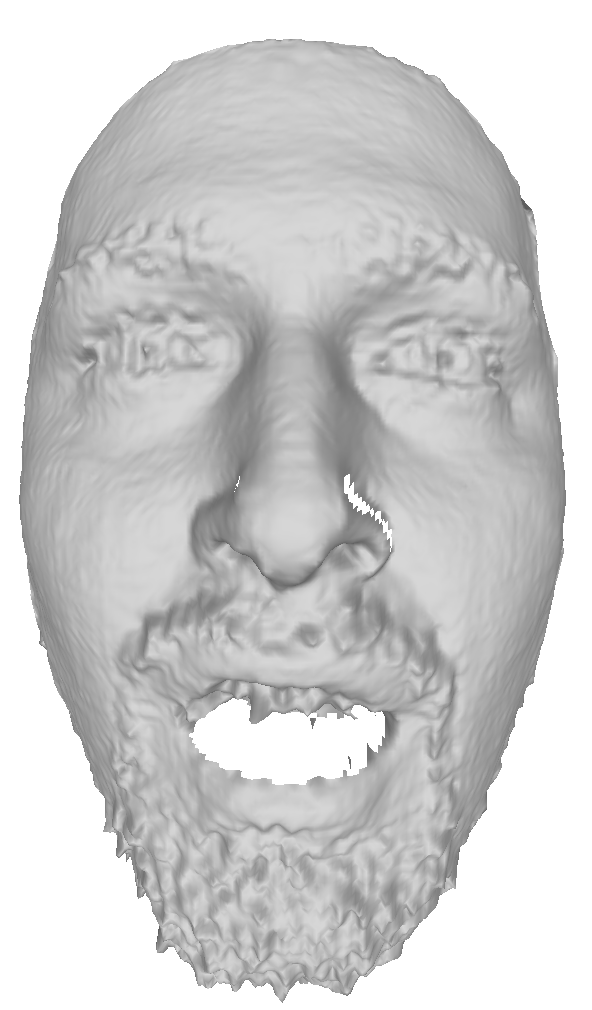}\\
\includegraphics[width=0.125\textwidth]{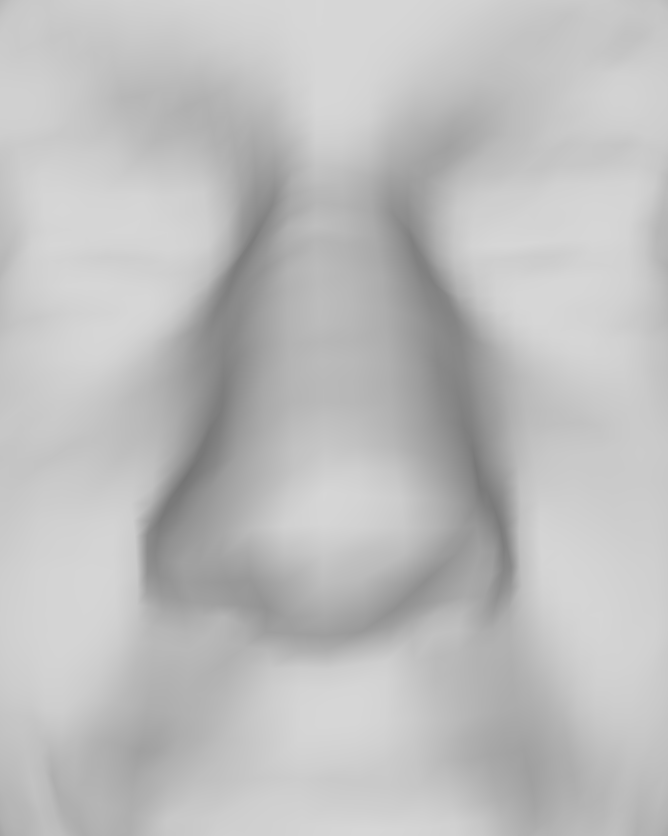}\hfill
\includegraphics[width=0.125\textwidth]{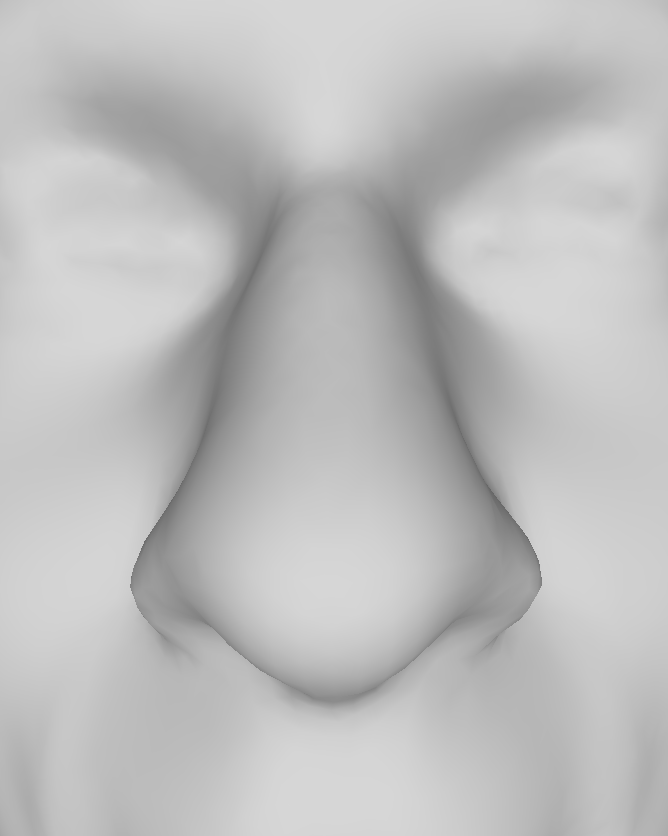}\hfill
\includegraphics[width=0.125\textwidth]{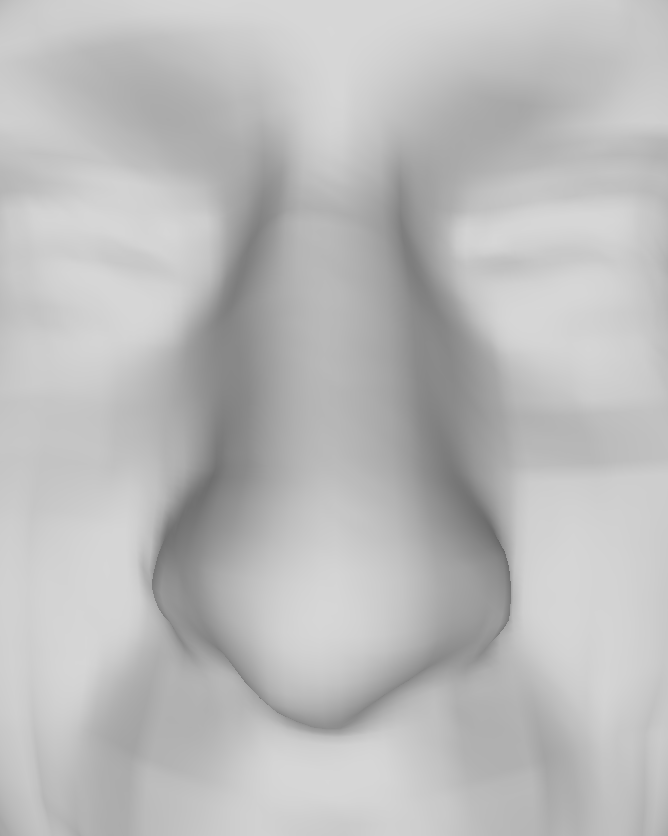}\hfill
\includegraphics[width=0.125\textwidth]{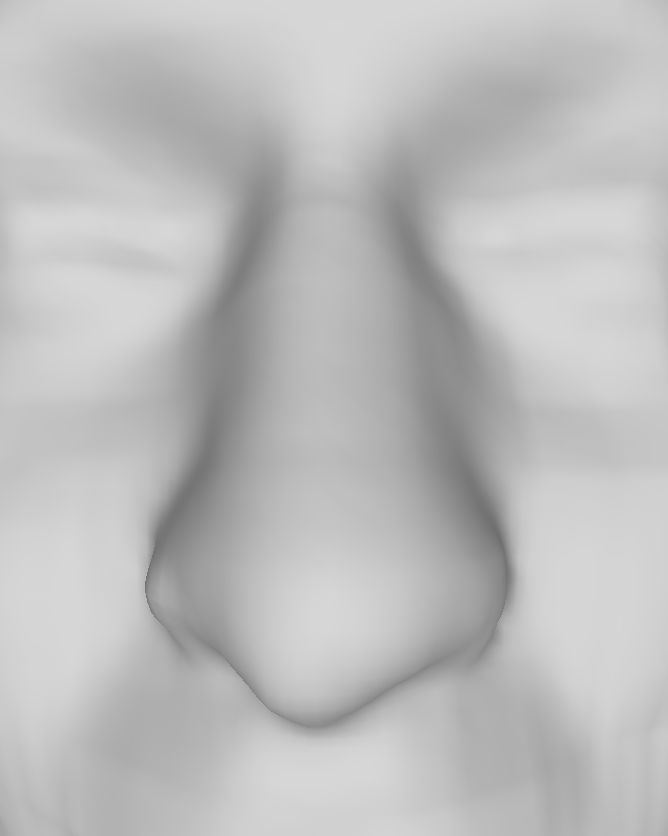}\hfill
\includegraphics[width=0.125\textwidth]{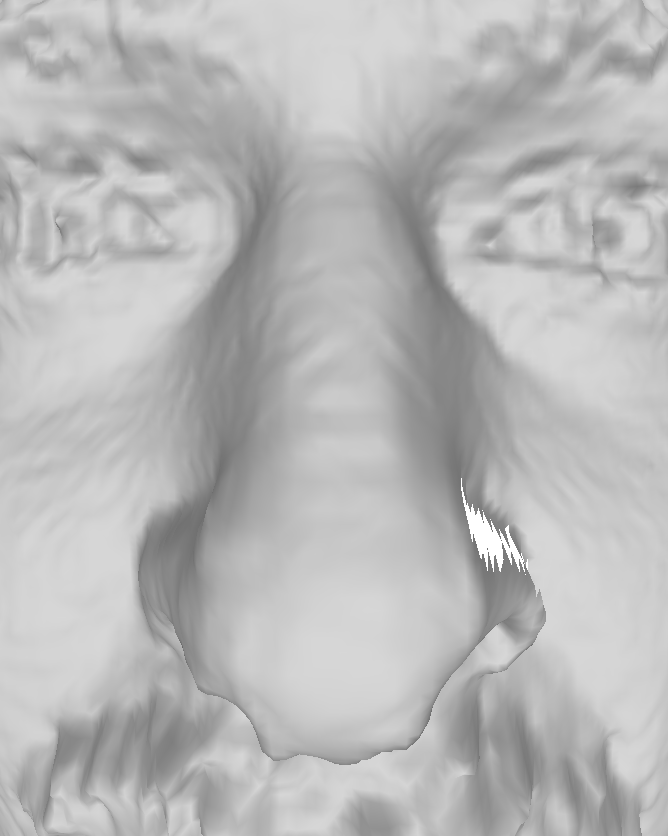}\\
}
\hfill
\parbox{0.30\textwidth}
{
\includegraphics[width=0.3\textwidth]{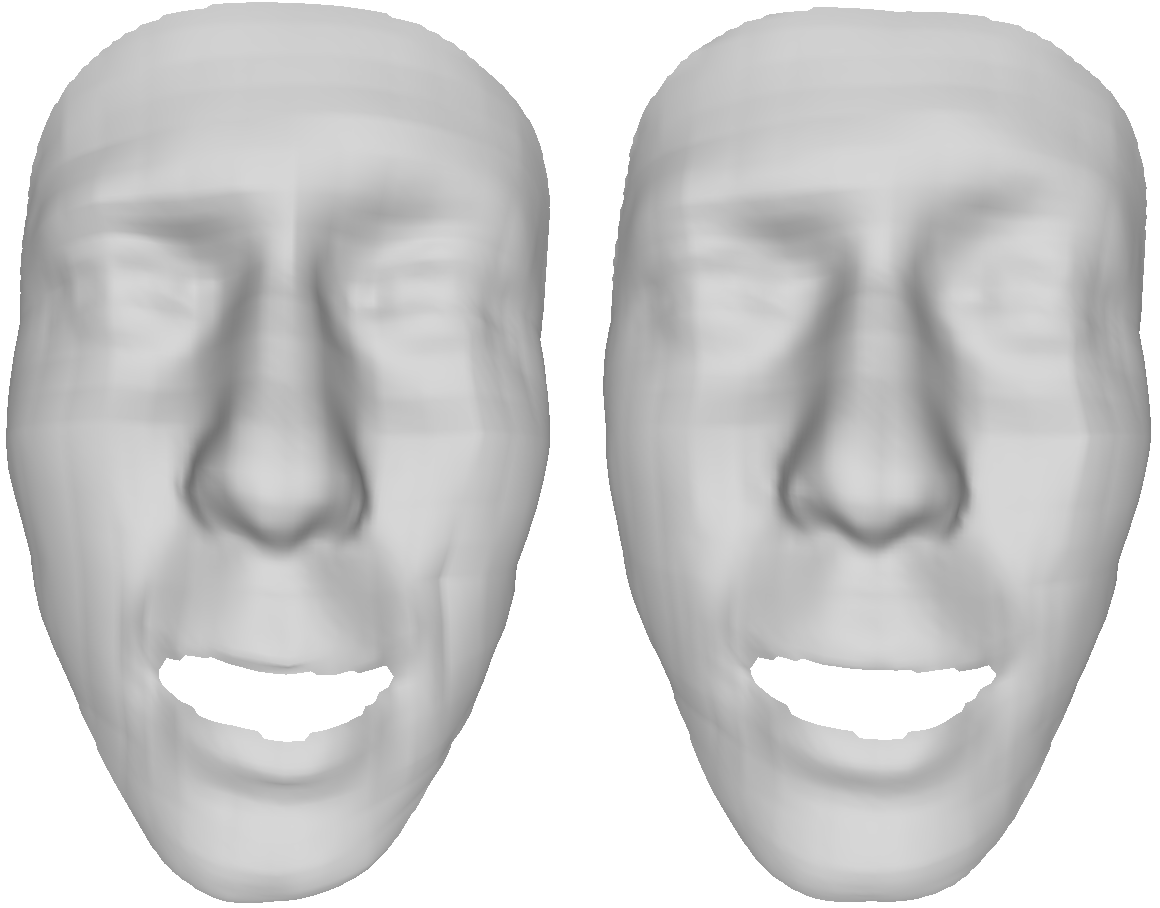}\hfill
}
\caption{Effect of smoothing energy $E_S$ on an example noisy scan. Left block: fitting results for a scan in surprise expression, with a close-up of the nose region in the bottom row. Left to right: local multiple PCA, global multilinear model, our model ($\weightfunc_S=0$), our model ($\weightfunc_S=100$), and input data. Right block: our reconstructions for a fear expression for $\weightfunc_S=0$ (left) and $\weightfunc_S=100$. Note the faint grid-artifacts that appear without smoothing, eg. in the cheek region and around the mouth. The input data can be seen in Figure~\ref{fig:NoisyDataEgs} (left block).}
\label{fig:smoothingDiff}
\end{figure*}

\begin{figure}[htp]
\parbox{0.475\textwidth}
{
\includegraphics[width=0.11\textwidth]{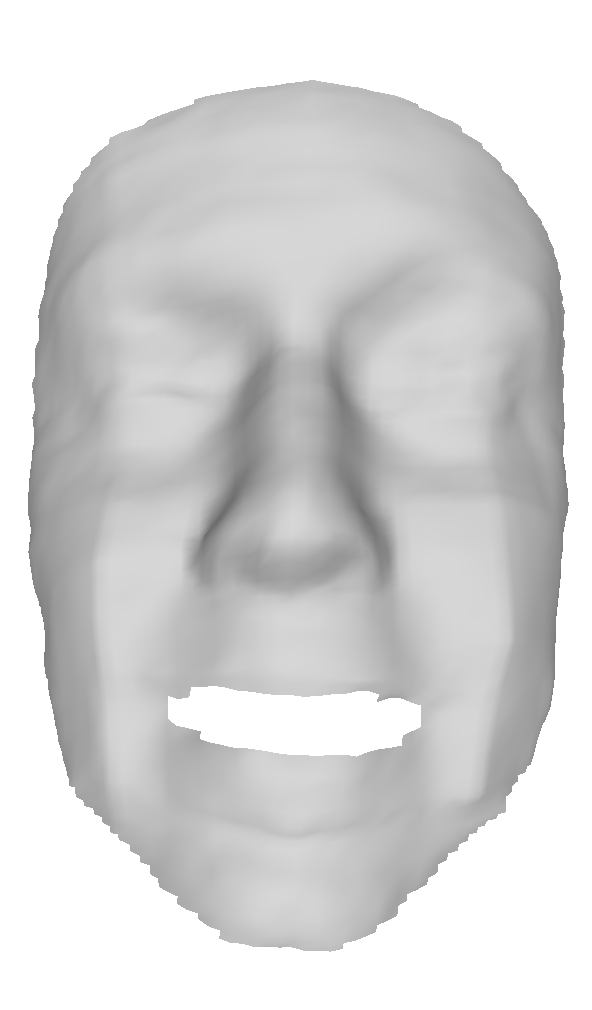}\hfill
\includegraphics[width=0.11\textwidth]{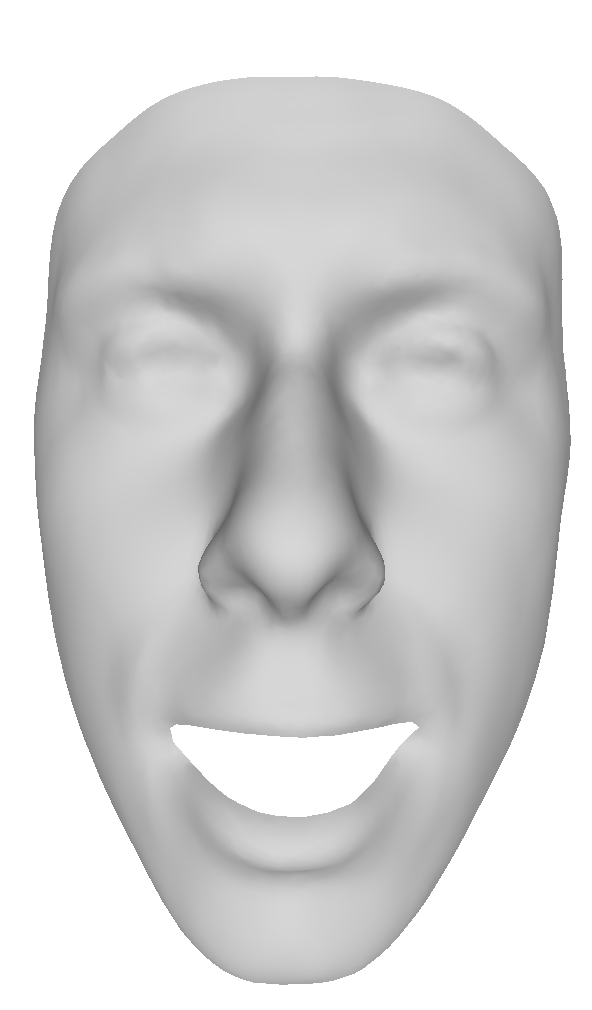}\hfill
\includegraphics[width=0.11\textwidth]{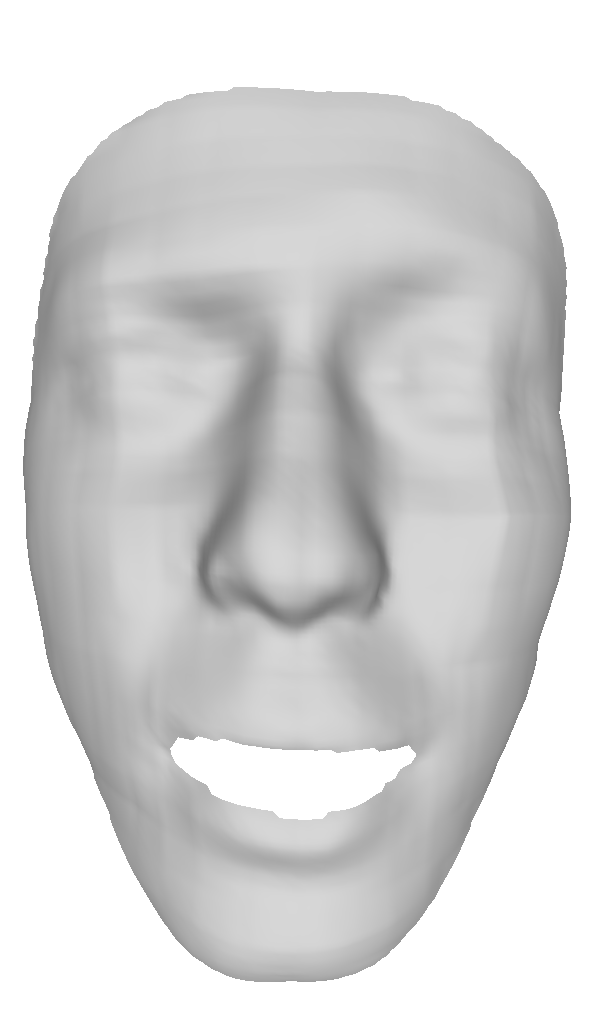}\hfill
\includegraphics[width=0.11\textwidth]{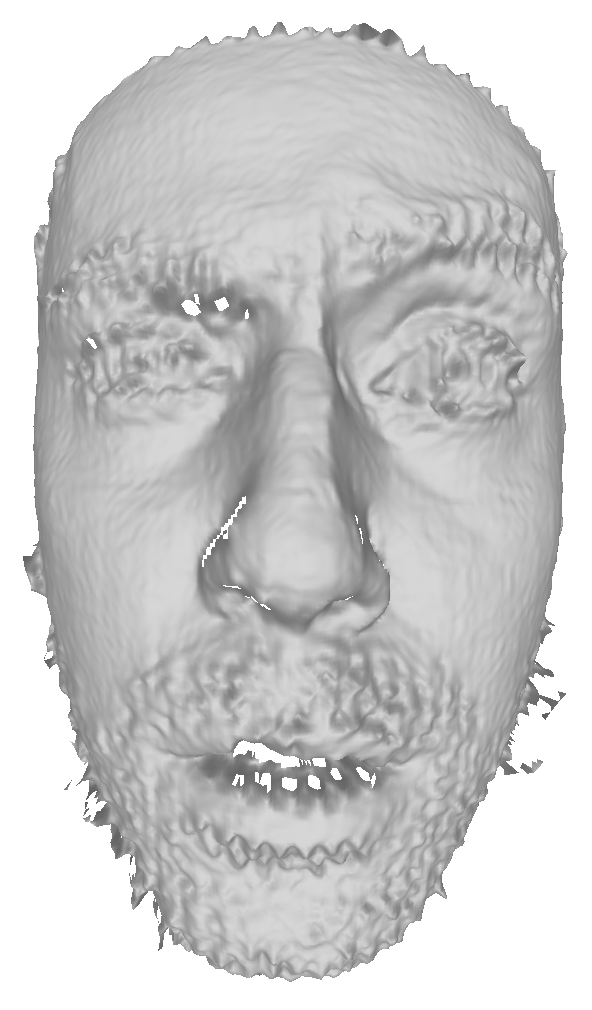}
}
\hfill
\parbox{0.475\textwidth}
{
\includegraphics[width=0.11\textwidth]{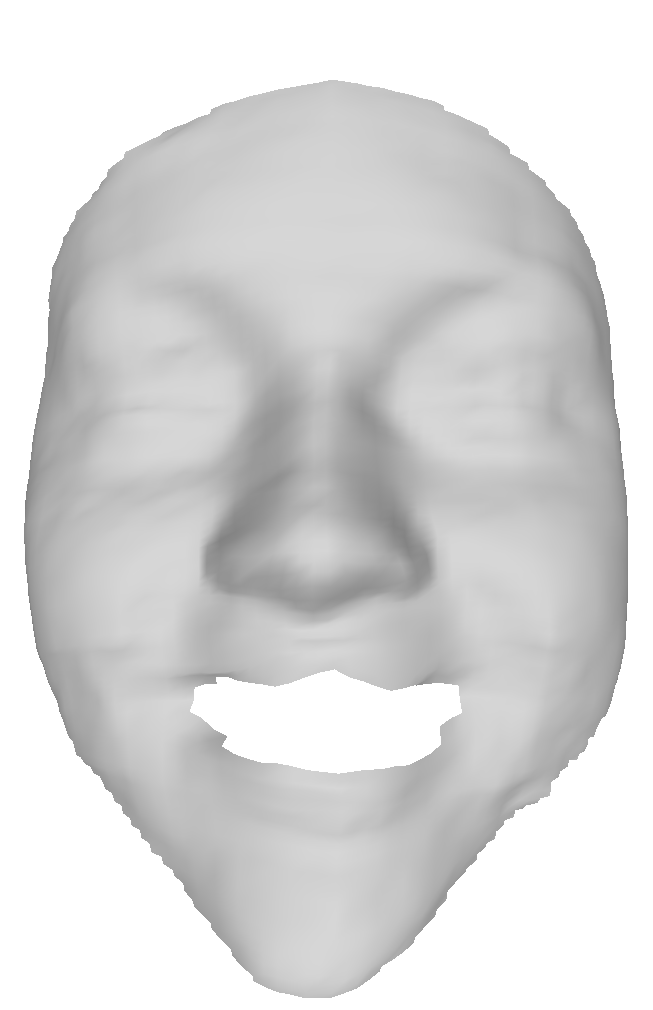}\hfill
\includegraphics[width=0.11\textwidth]{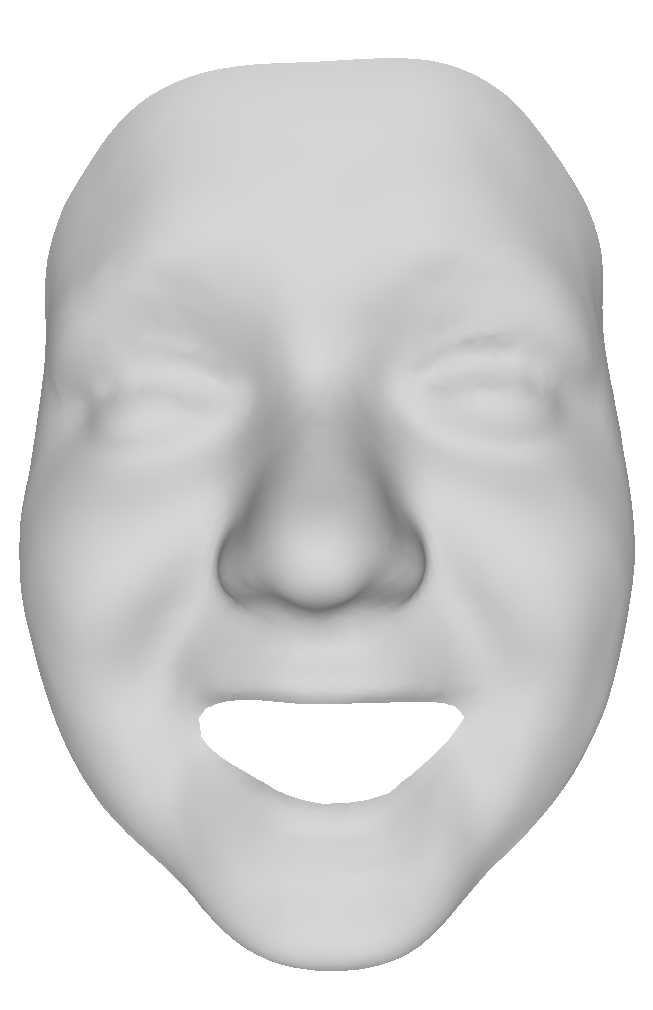}\hfill
\includegraphics[width=0.11\textwidth]{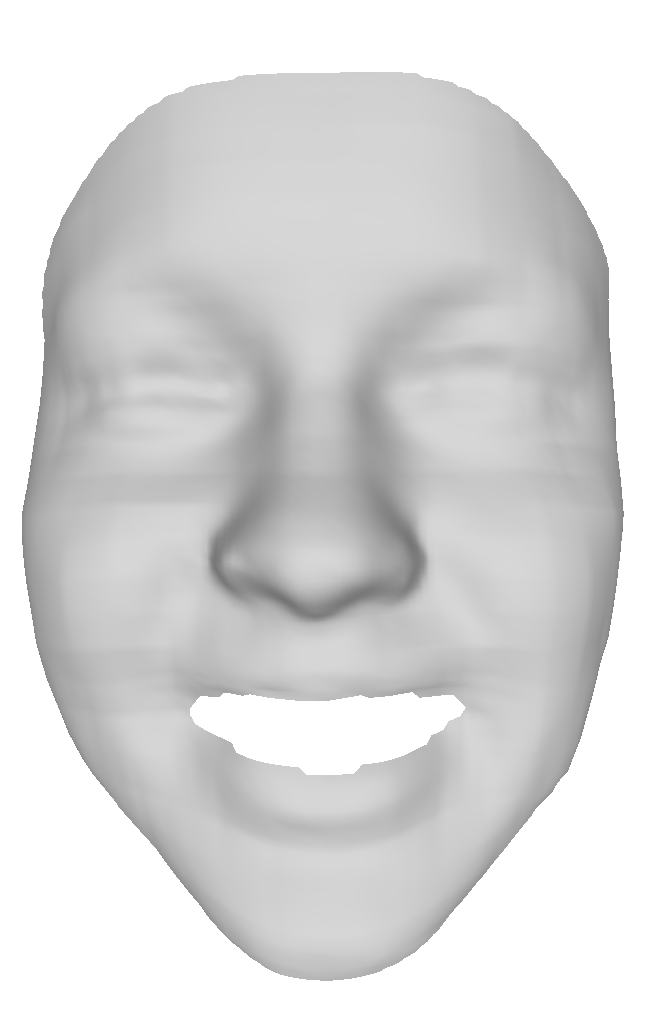}\hfill
\includegraphics[width=0.11\textwidth]{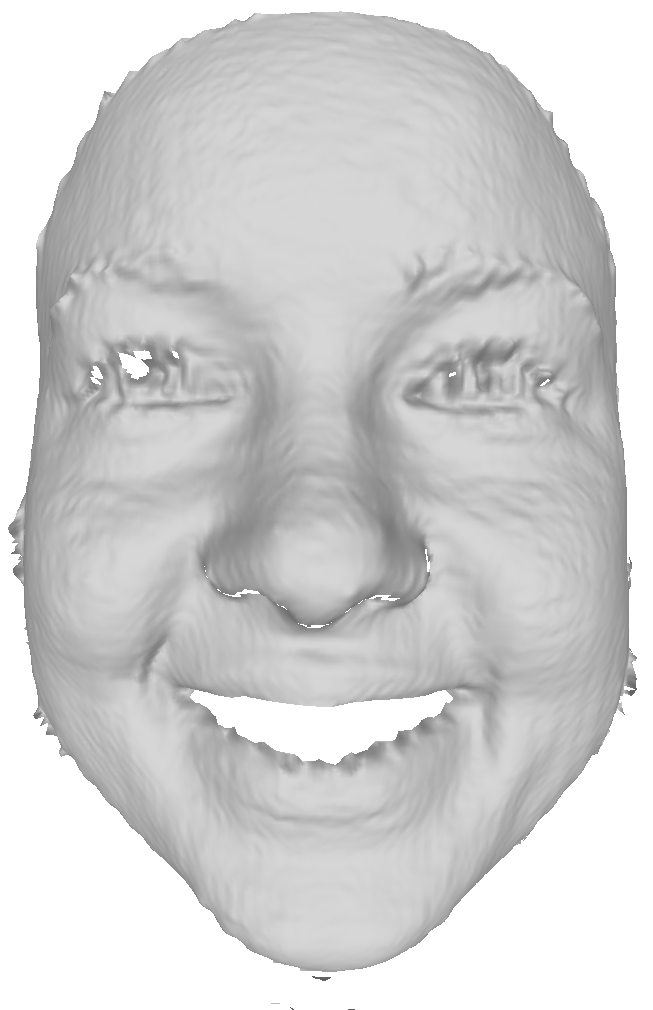}
}
\caption{Reconstruction examples for noisy scans in different expressions. 
Top block: fear expression. Top block: happy expression. Each block, from left to right: local multiple PCA~\cite{Brunton2011}, global multilinear~\cite{BolkartWuhrer2013}, proposed ($\weightfunc_S=100$), input data.}
\label{fig:NoisyDataEgs}
\end{figure}

In this section, we demonstrate our model's ability to capture fine-scale detail in the presence of identity and expression variation, and high noise levels. We fit it to $120$ models ($20$ identities in up to $7$ expressions) from the Bosphorus database~\cite{Bosphorus_2008}. We measure the fitting error as distance-to-data, and the per-vertex median errors are shown for all three models in Figure \ref{fig:NoisyDataQuant} (left). Our model has a greater proportion of sub-millimeter errors than either of the other models. Specifically, the local PCA and the global multilinear have $63.2\%$ and $62.0\%$, respectively, of vertices with error $<1\mbox{mm}$, whereas our model has $71.6\%$ with $\weightfunc_S=100$ and $72.4\%$ with $\weightfunc_S=0$. Figure \ref{fig:NoisyDataQuant} (right) shows cumulative error plots for all three methods for vertices in the characteristic detail region of the face, which is shown next to the plot. This region contains prominent facial features with the most geometric detail. We see that our model is more accurate than previous models in this region and has many more sub-millimeter errors; the local PCA and global multilinear have $60.4\%$ and $58.0\%$ of errors $<1\mbox{mm}$, respectively, whereas our model has $70.2\%$ with $\weightfunc_S=100$ and $72.7\%$ with $\weightfunc_S=0$. This shows that our model has improved accuracy for fine-scale detail compared to existing models, in particular in areas with prominent features and high geometric detail. 

Figures \ref{fig:smoothingDiff} and \ref{fig:NoisyDataEgs} show examples of fitting to noisy scans of different subjects in different expressions. These scans contain acquisition noise, missing data and facial hair. Figure \ref{fig:smoothingDiff} (left) shows a surprise expression and close-ups of the nose region; our reconstruction both $\weightfunc_S=100$ and $\weightfunc_S=0$ capture significantly more fine-scale detail than previous models. The right part of the figure demonstrates the effect of the smoothing energy in preventing faint grid artifacts appearing in the reconstruction due to the independent optimization scheme. Figure \ref{fig:NoisyDataEgs} shows two subjects in fear and happy expressions. We again see the increased accuracy of our model in terms of fine-scale detail on facial features compared to previous models. Note the accuracy of the nose and mouth shapes in all examples compared to the other models, and the accurate fitting of the underlying face shape in the presence of facial hair. Further note how our model captures the asymmetry in the eyebrow region for the fear expression.

\subsection{Reconstruction of Occluded Data}
\label{sec_eval_occluded}

\begin{figure}[t]
\centering
\parbox{0.45\textwidth}
{
\raisebox{1.8cm}{
\parbox{0.185\textwidth}{
\includegraphics[width=0.09\textwidth]{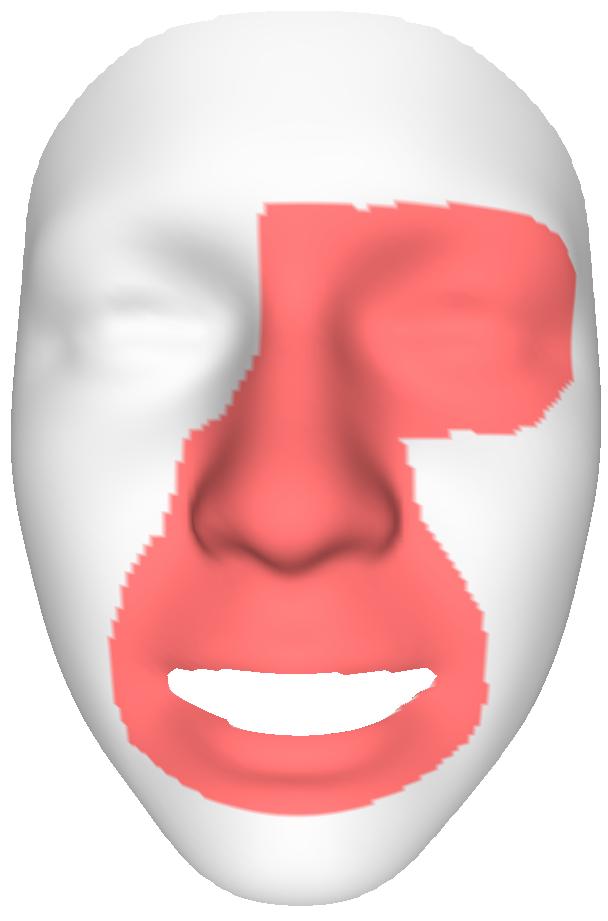}\hfill
\includegraphics[width=0.09\textwidth]{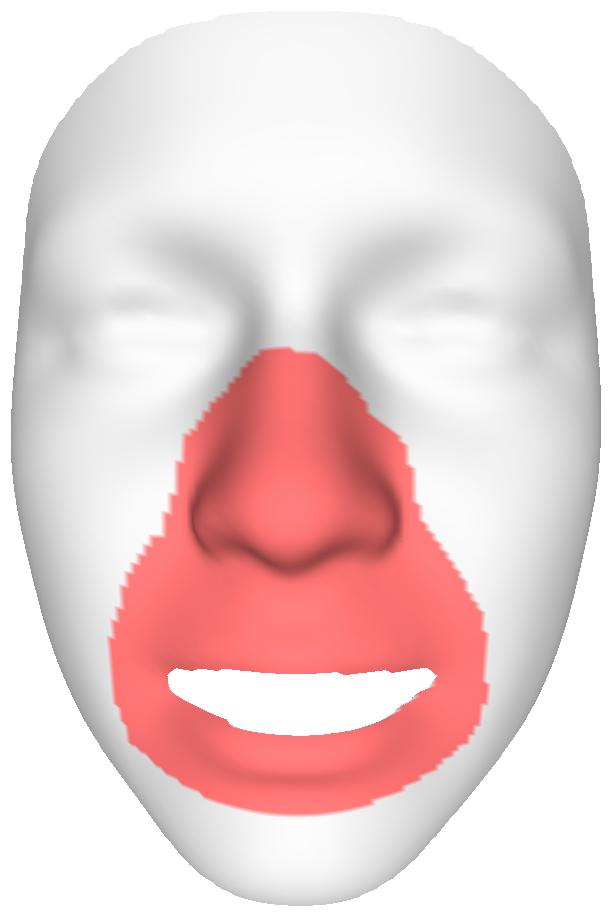}\\
\includegraphics[width=0.09\textwidth]{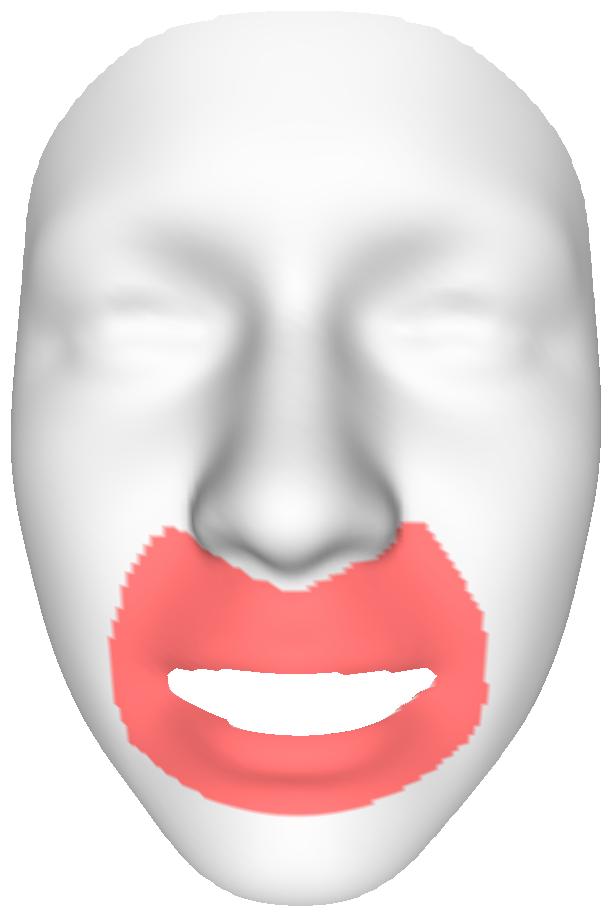}\hfill
\includegraphics[width=0.09\textwidth]{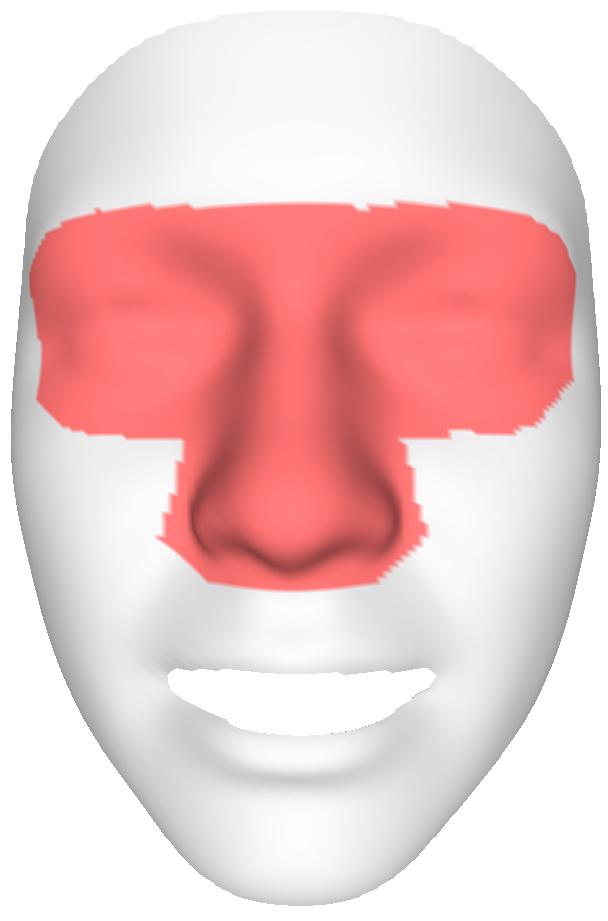}
}
}
\hfill
\raisebox{0.3cm}{
\includegraphics[width=0.225\textwidth]{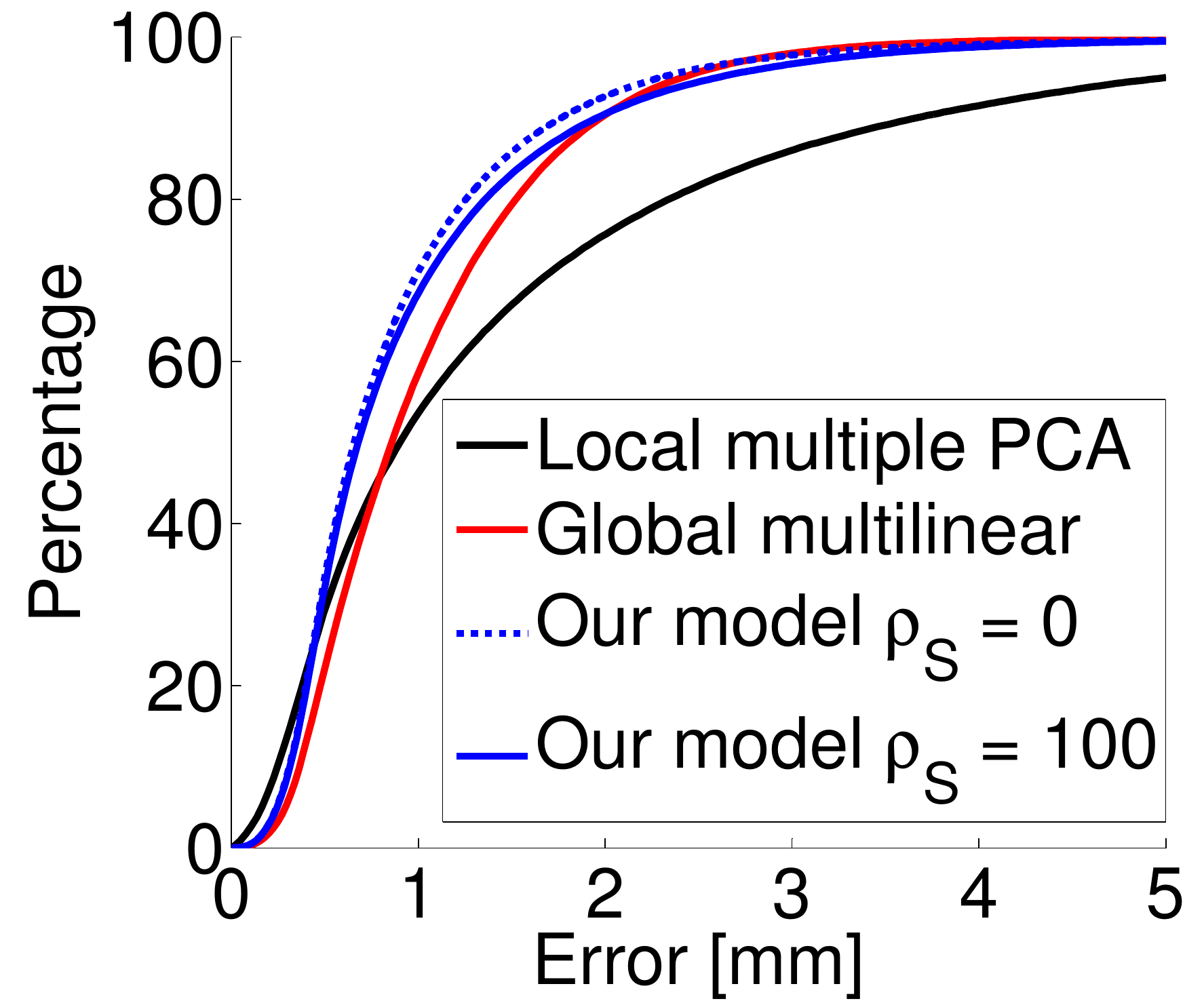}
}
}
\\
\parbox{0.45\textwidth}
{
\includegraphics[width=0.11\textwidth]{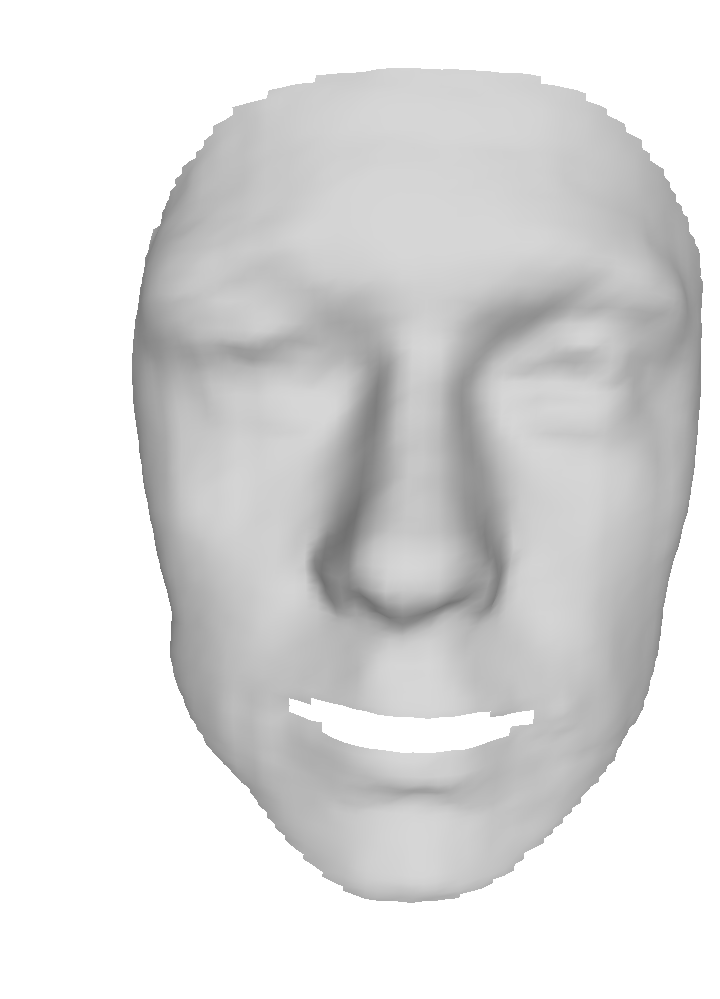}\hfill
\includegraphics[width=0.11\textwidth]{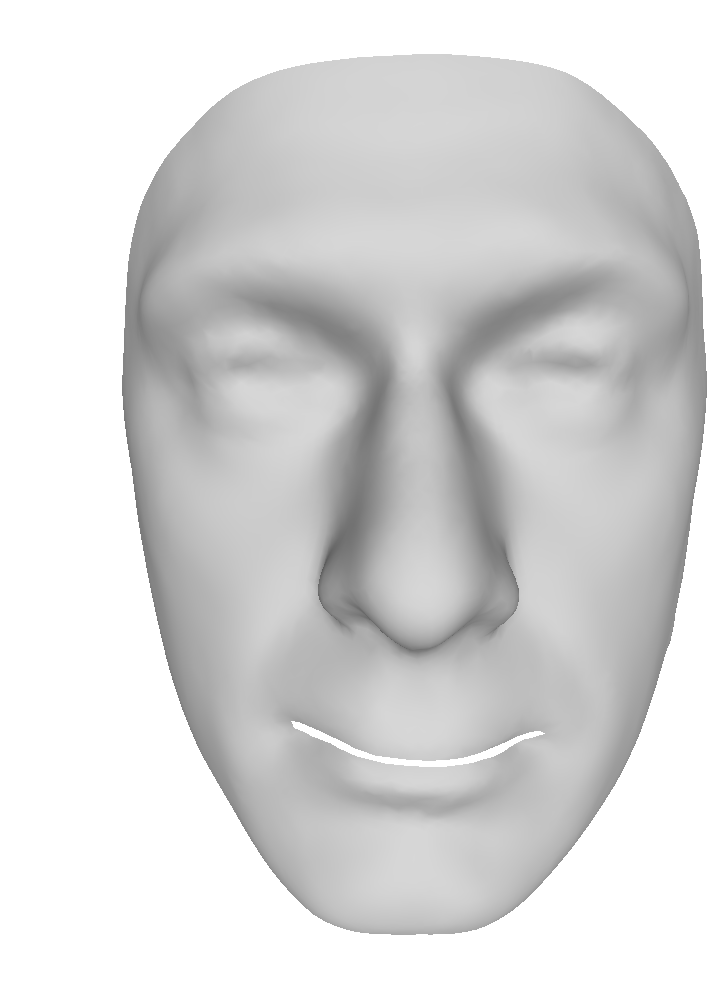}\hfill
\includegraphics[width=0.11\textwidth]{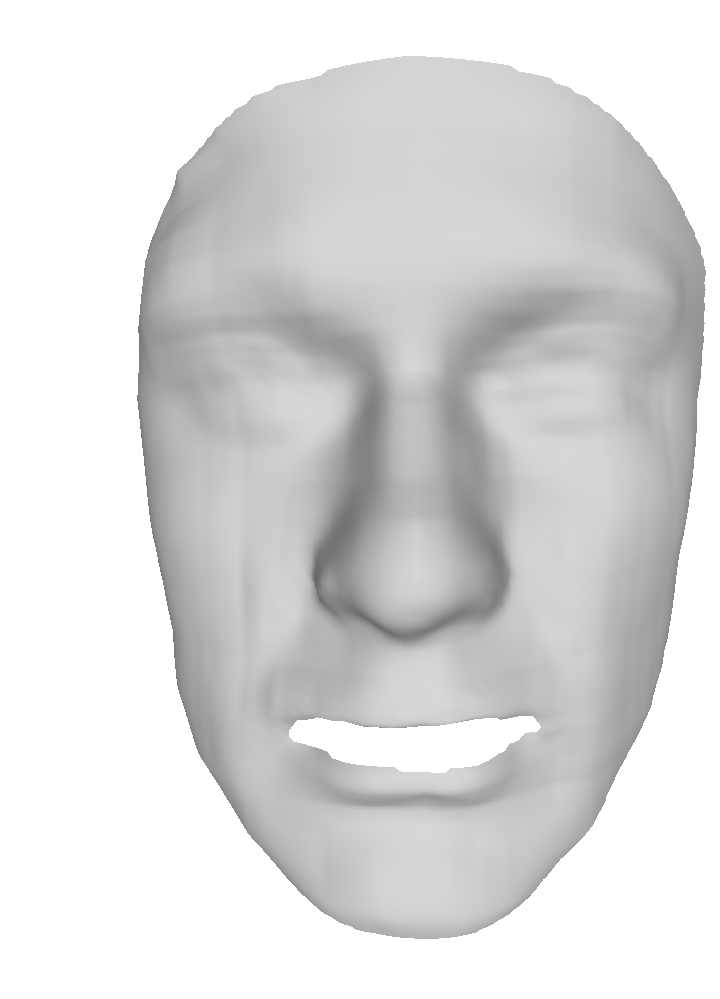}\hfill
\includegraphics[width=0.11\textwidth]{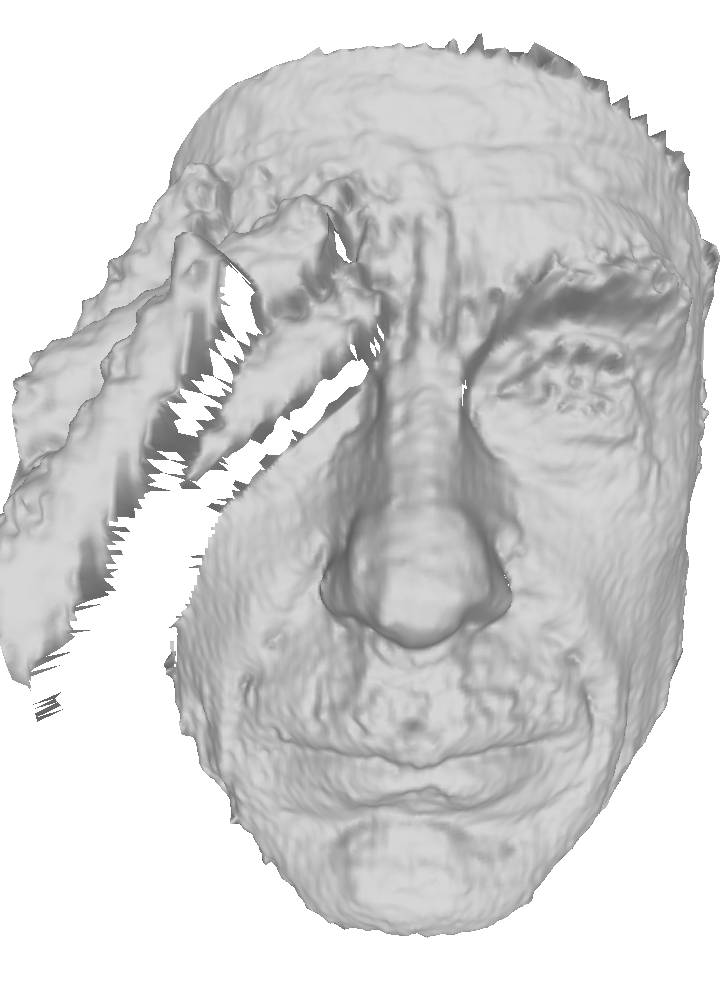}\\
\includegraphics[width=0.11\textwidth]{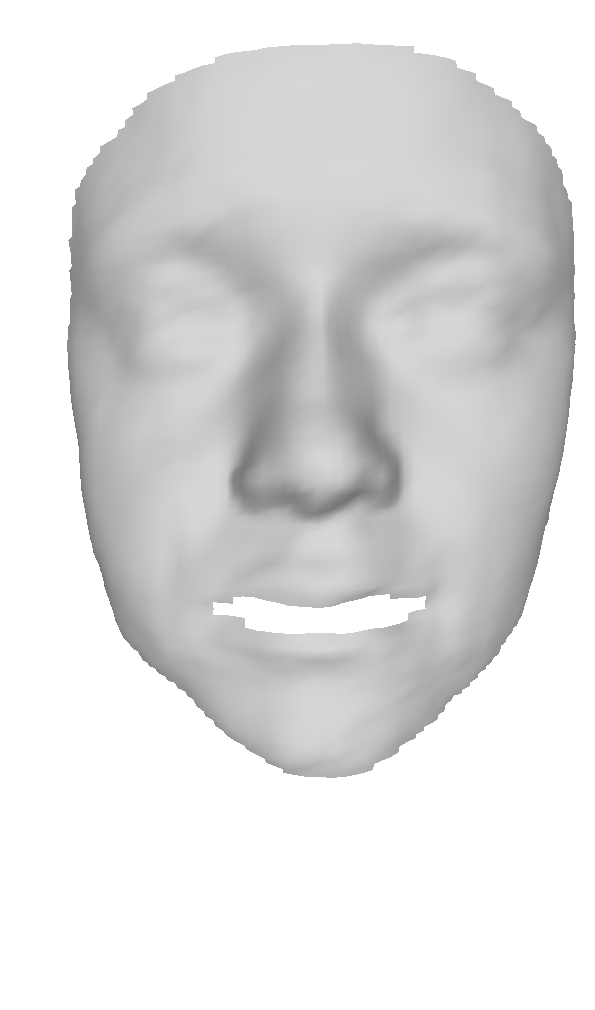}\hfill
\includegraphics[width=0.11\textwidth]{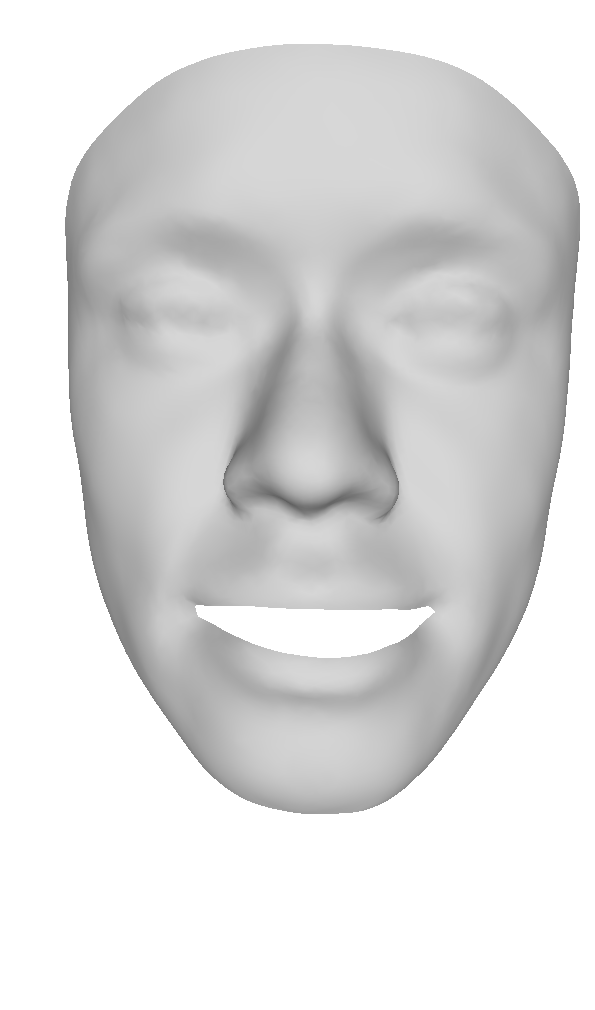}\hfill
\includegraphics[width=0.11\textwidth]{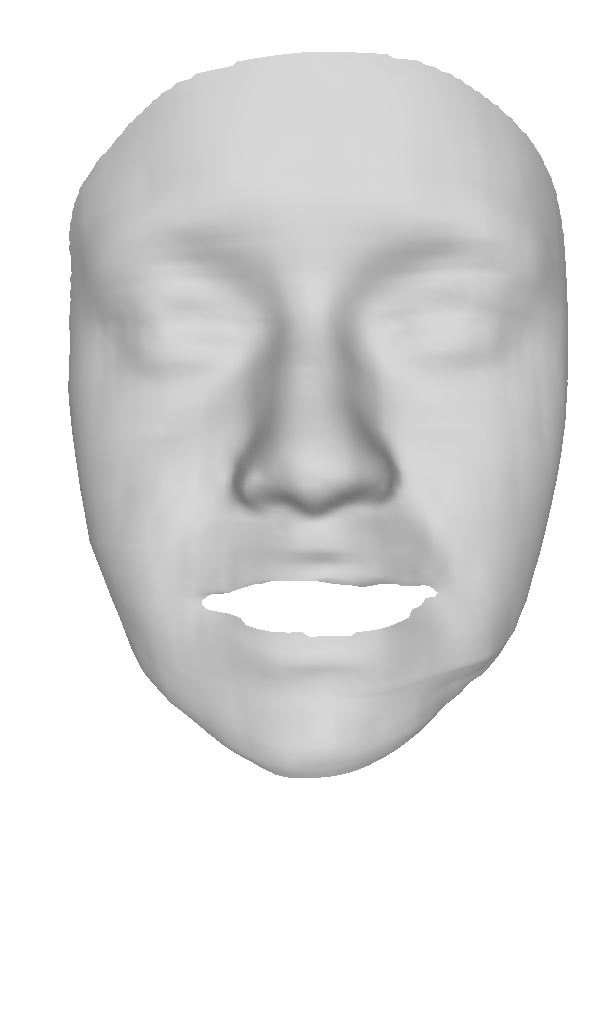}\hfill
\includegraphics[width=0.11\textwidth]{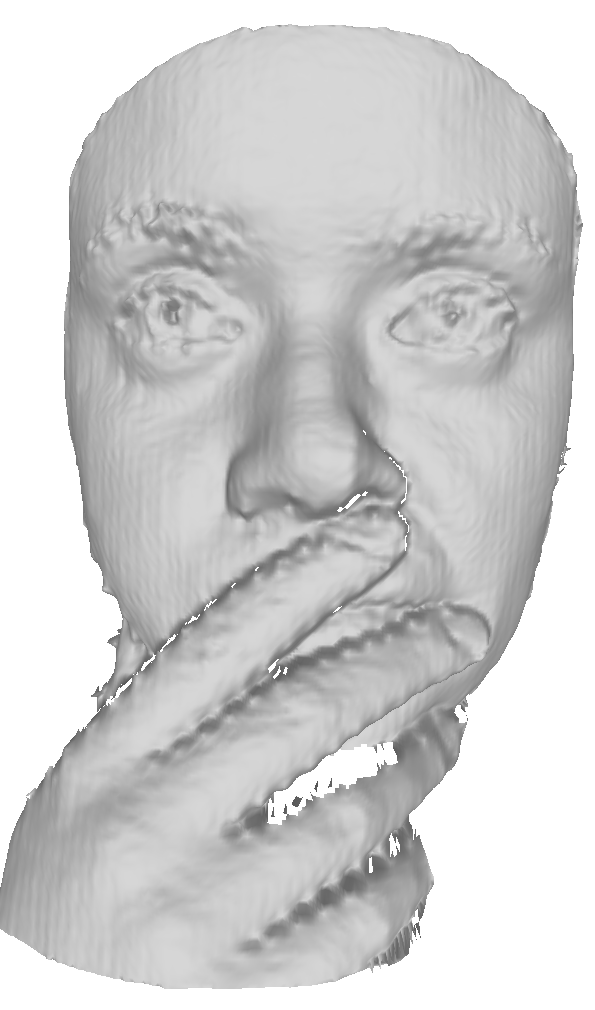}
}
\caption{Top left: Masks used to measure error for the different occlusions types. Top right: combined cumulative error plot. Bottom two rows: reconstruction examples for a scans with occlusions (eye and mouth). Each row: local multiple PCA model, global multilinear model, our reconstruction with $\weightfunc_S=100$, input data.}
\label{fig:OcclData}
\end{figure}

In this section, we demonstrate our model's robustness to severe data corruptions in the form of occlusions. We fit all three models to $80$ scans ($20$ subjects, $4$ types of occlusions) from the Bosphorus database~\cite{Bosphorus_2008}. Figure \ref{fig:OcclData} (top right) shows the cumulative error for all three models. Since distance-to-data is not a valid error measure in occluded areas, we apply different masks, shown next to the error plot, depending on the type of occlusion so that only unoccluded vertices are measured. Clockwise from top-left: the mask used for eye, glasses, mouth and hair occlusions. From the cumulative error curves, we see that our model retains greater accuracy in unoccluded parts of the face than previous models.

The bottom two rows of Figure \ref{fig:OcclData} show example reconstructions in the presence of severe occlusions. All models show robustness to occlusions and reconstruct plausible face shapes, but our model provides better detail in unoccluded parts of the face than previous models (see the mouth and chin in the first row, and the nose in the second row). For these examples, we show our reconstruction with $\weightfunc_S=100$.

\subsection{Reconstruction of Motion Data}
\label{sec_eval_motion}

\begin{figure*}[t]
\centering
\parbox{0.65\textwidth}
{
\includegraphics[width=0.15\textwidth]{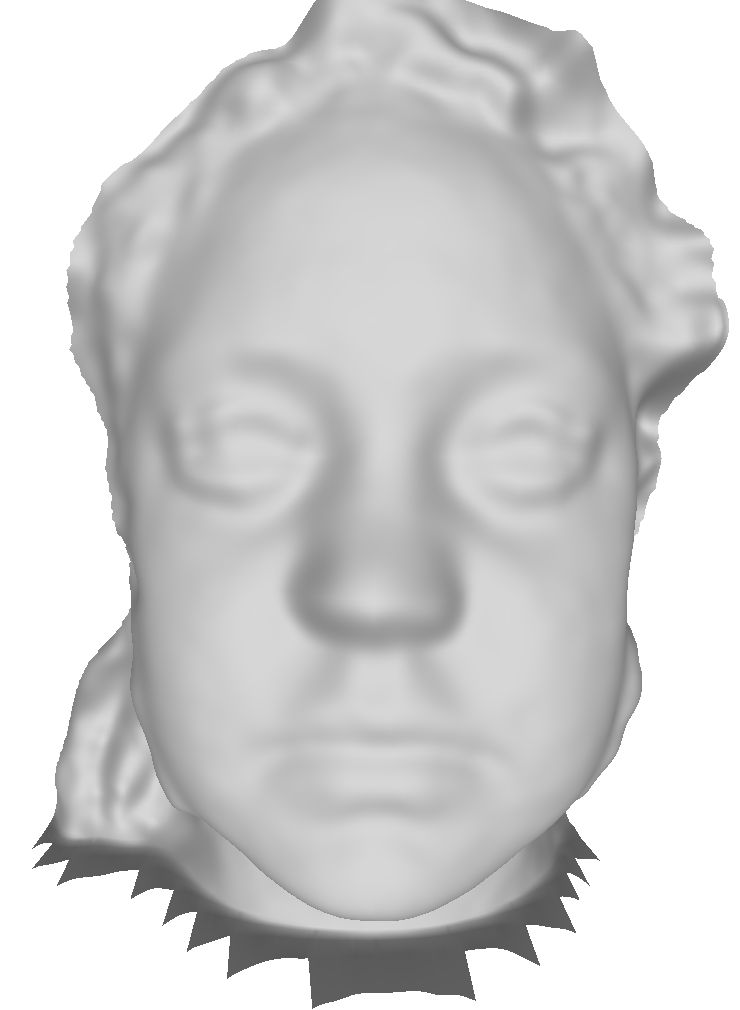}\hfill
\includegraphics[width=0.15\textwidth]{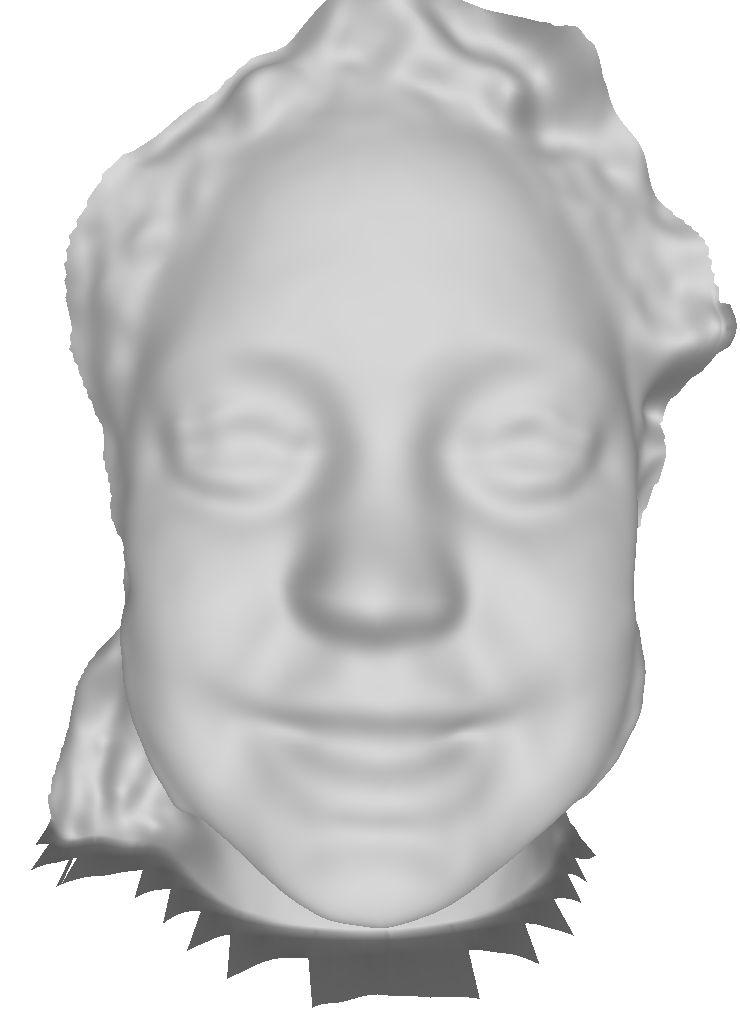}\hfill
\includegraphics[width=0.15\textwidth]{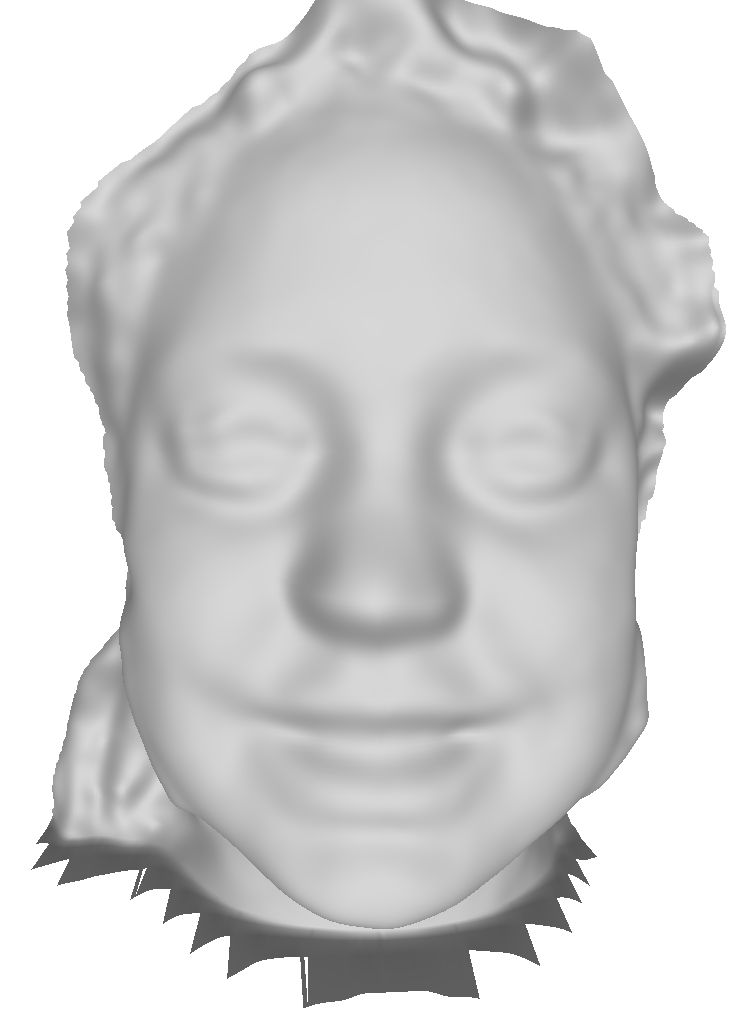}\hfill
\includegraphics[width=0.15\textwidth]{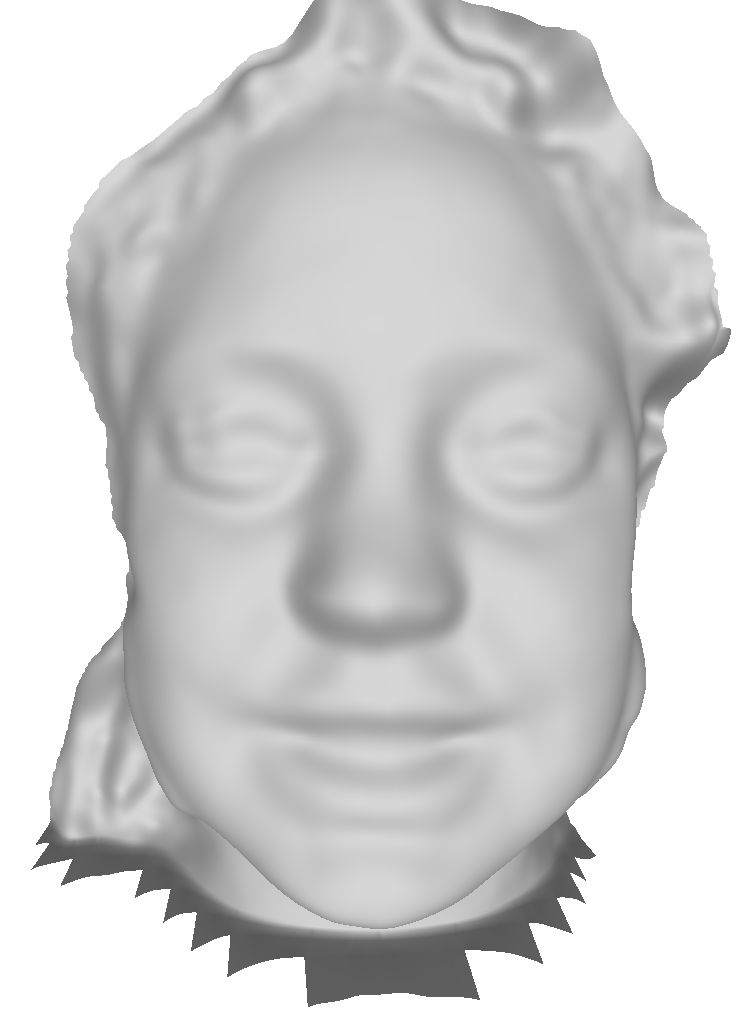}\\
\includegraphics[width=0.15\textwidth]{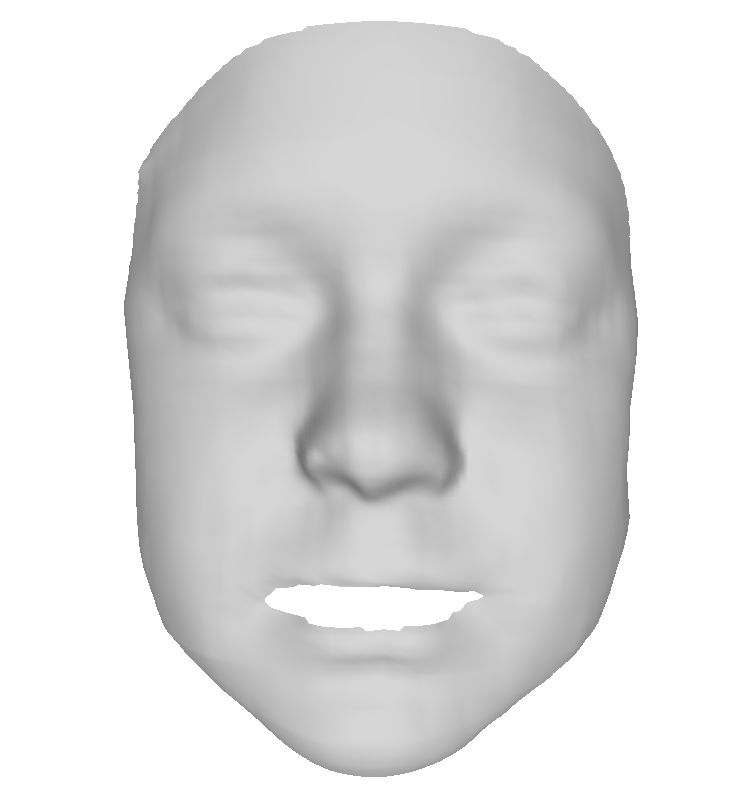}\hfill
\includegraphics[width=0.15\textwidth]{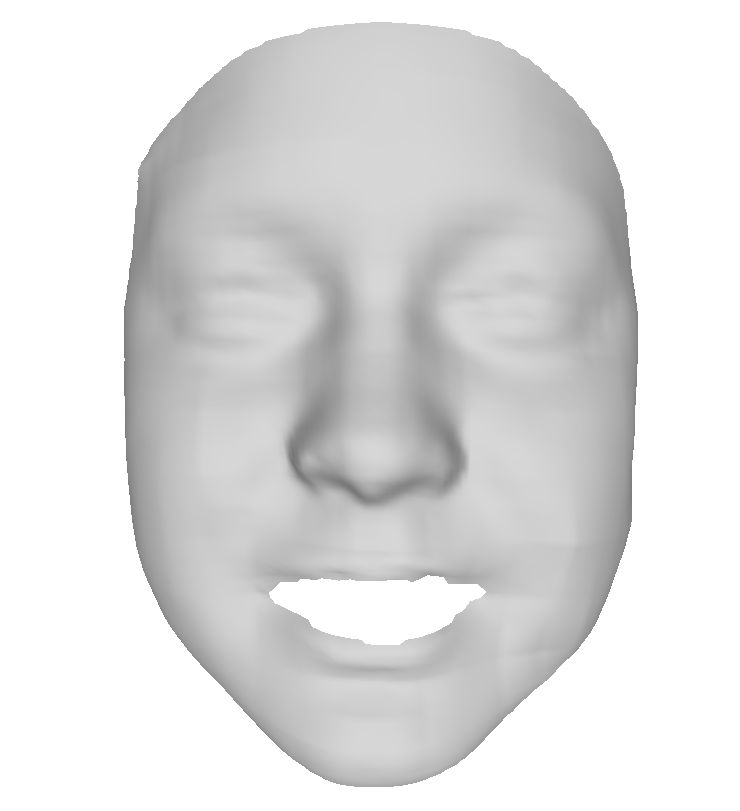}\hfill
\includegraphics[width=0.15\textwidth]{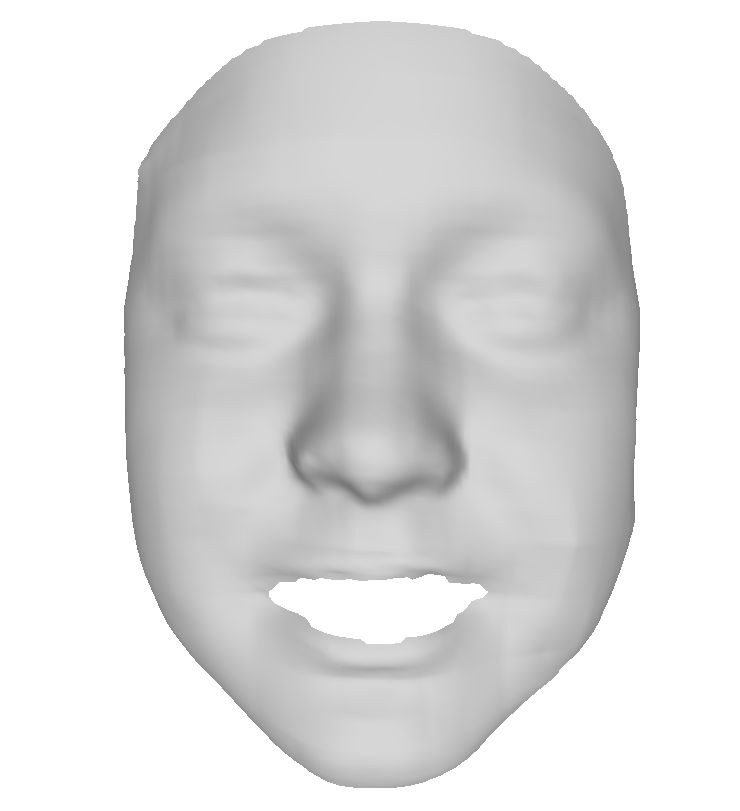}\hfill
\includegraphics[width=0.15\textwidth]{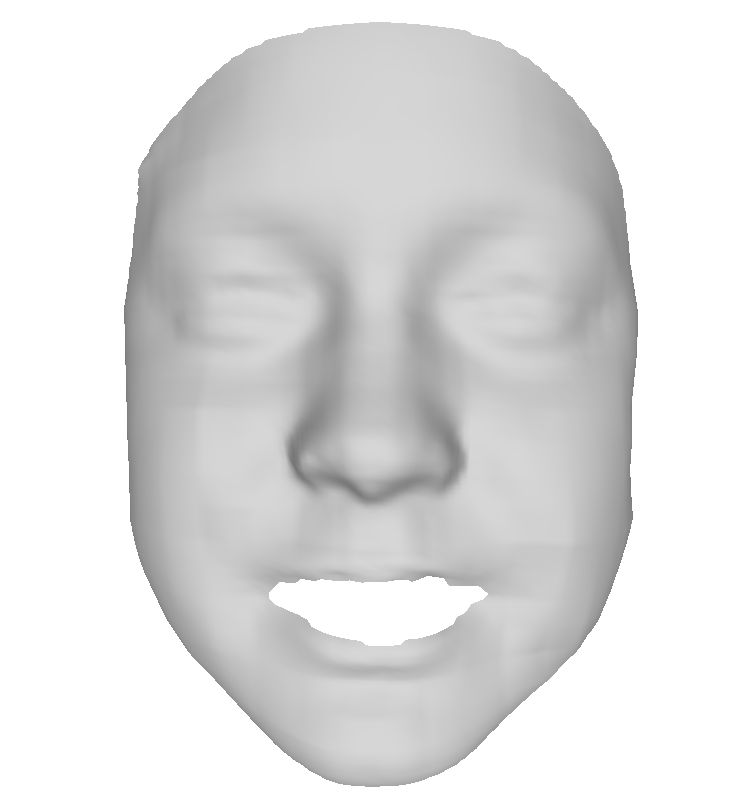}\\
}
\\
\parbox{0.65\textwidth}
{
\includegraphics[width=0.15\textwidth]{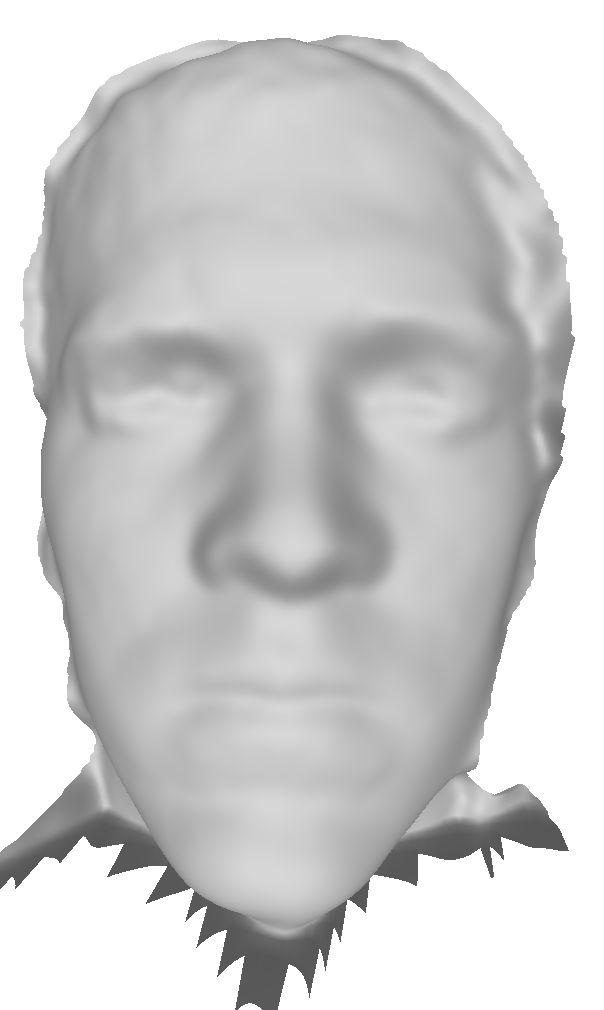}\hfill
\includegraphics[width=0.15\textwidth]{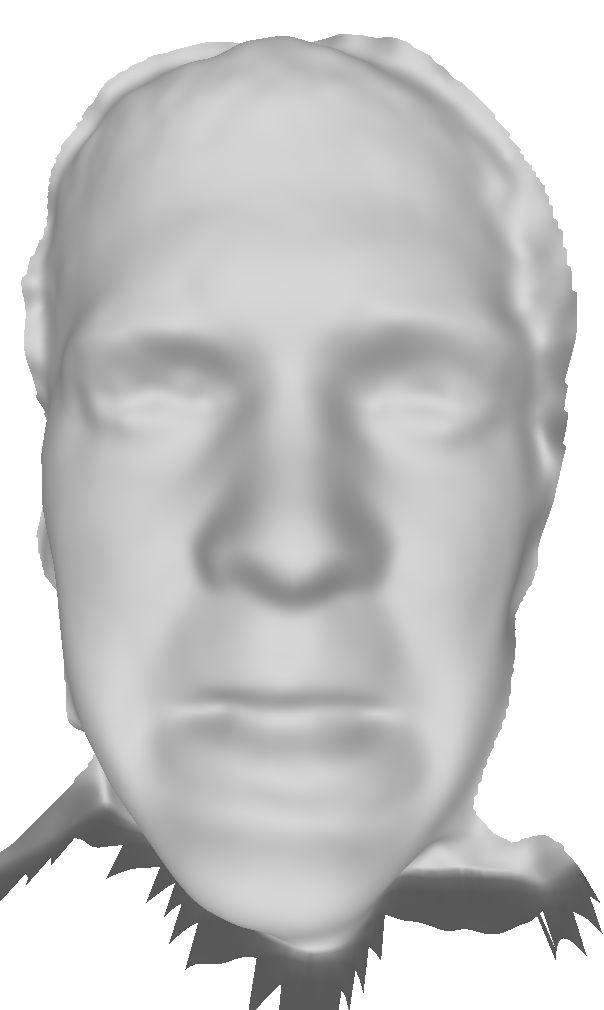}\hfill
\includegraphics[width=0.15\textwidth]{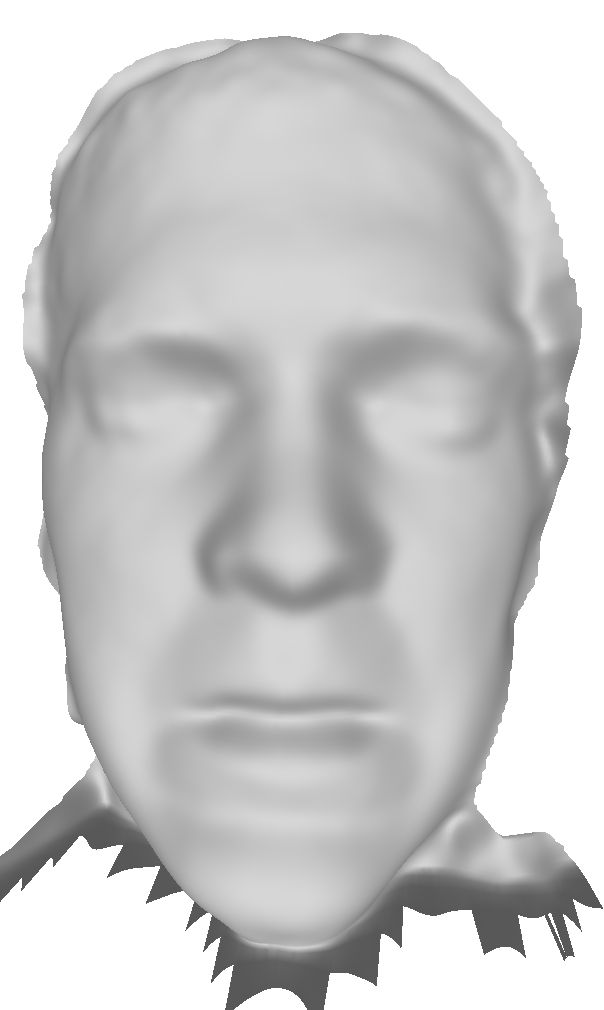}\hfill
\includegraphics[width=0.15\textwidth]{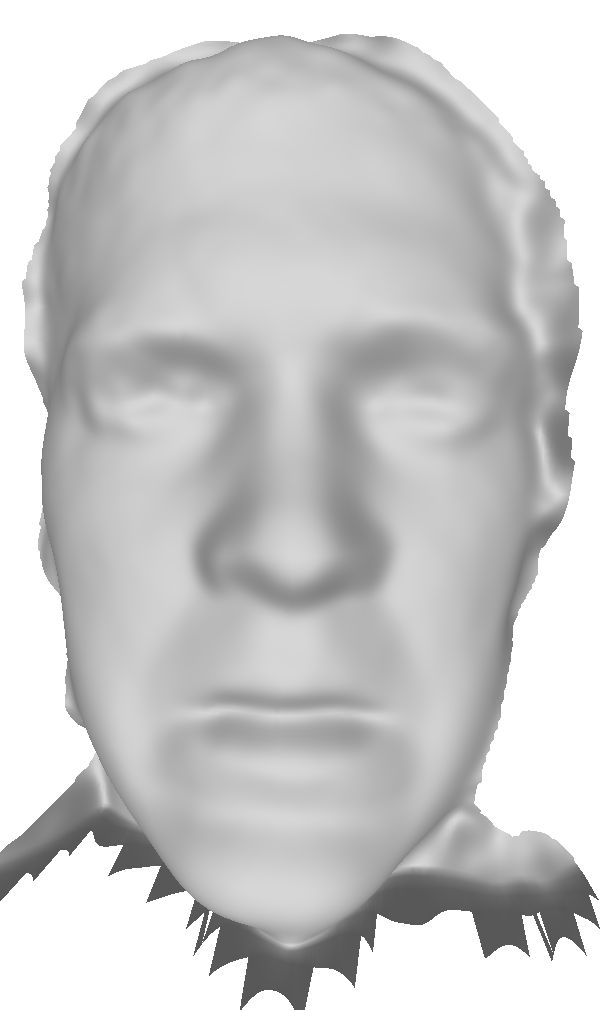}\\
\includegraphics[width=0.15\textwidth]{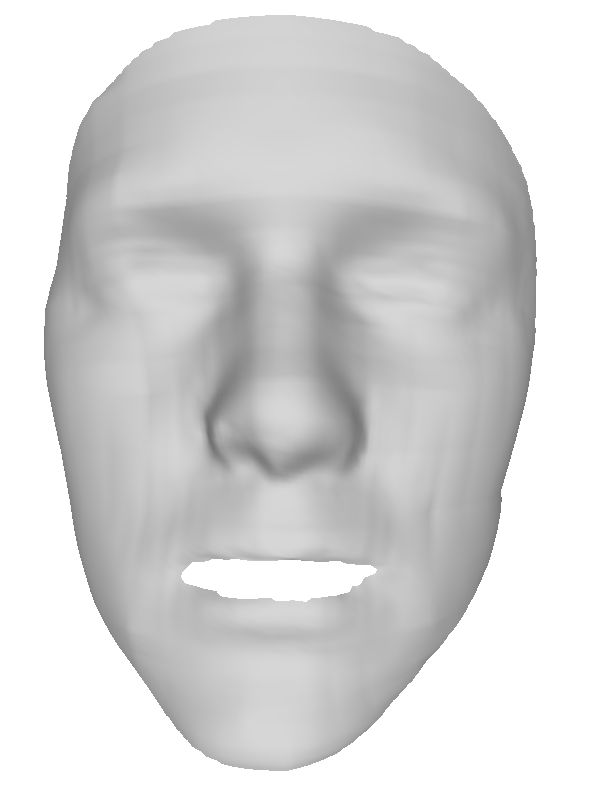}\hfill
\includegraphics[width=0.15\textwidth]{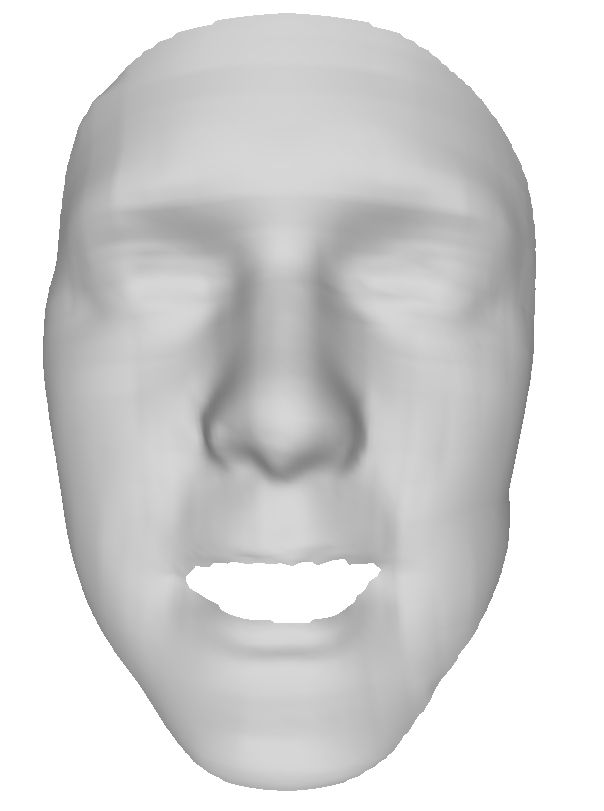}\hfill
\includegraphics[width=0.15\textwidth]{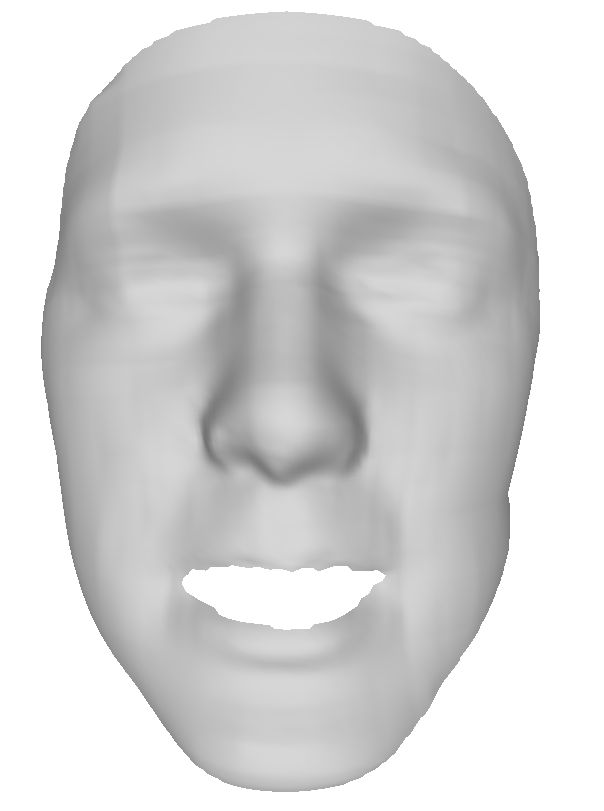}\hfill
\includegraphics[width=0.15\textwidth]{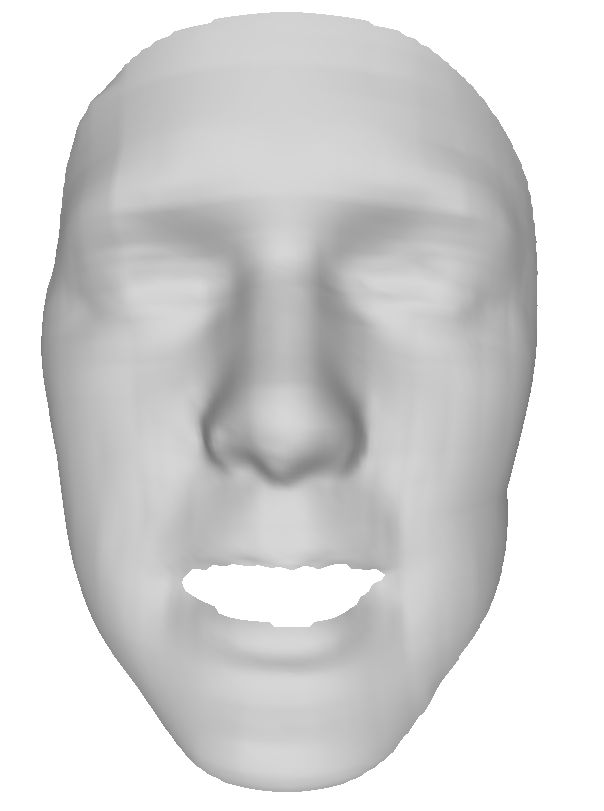}\\
}
\caption{Tracking results for the application of our fitting algorithm given in Section \ref{sec_fitting_track}. Each block shows frames 0, 20, 40 and 60 of a sequence of a subject performing an expression. Top: happy expression. Bottom: fear expression.}
\label{fig:MotionData}
\end{figure*}

In this section, we show our model's applicability to $3$D face tracking using the simple extension to our fitting algorithm described in Section \ref{sec_fitting_track}. Figure \ref{fig:MotionData} shows some results for a selection of frames from three sequences from the BU4DFE database~\cite{BU-4DFE_2008}. We see that, as for static scans, high levels of facial detail are obtained, and even the simple extension of our fitting algorithm tracks the expression well. Since landmarks are predicted automatically for these sequences, the entire tracking is done automatically. This simple tracking algorithm is surprisingly stable. Videos can be found in the supplemental material.

\section{Conclusion}

We have presented a novel statistical shape space for human faces. Our multilinear wavelet model allows for reconstruction of fine-scale detail, while remaining robust to noise and severe data corruptions such as occlusions, and is highly efficient and scalable. The use of the wavelet transform has both statistical and computational advantages. By decomposing the surfaces into decorrelated wavelet coefficients, we can learn many independent low-dimensional statistical models rather than a single high-dimensional model. Lower dimensional models reduce the risk of overfitting, which allows us to set tight statistical bounds on the shape parameters, thereby providing robustness to data corruptions while capturing fine-scale detail. Model dimensionality is the dominant factor in the numerical routines used for fitting the model to noisy input data, and fitting many low-dimensional models is much faster than a single high-dimensional model even when the total number of parameters is much greater. We have demonstrated these properties experimentally with a thorough evaluation on noisy data with varying expression, occlusions and missing data. We have further shown how our fitting procedure can be easily and simply extended to give stable tracking of $3$D facial motion sequences. Future work includes making our model applicable for real-time tracking. Virtually all aspects of our fitting algorithm are directly parallelizable, and an optimized GPU implementation could likely achieve real-time fitting rates, in particular for tracking, where only expression weights need to be optimized every frame. Such high-detail real-time tracking could have tremendous impact in tele-presence and gaming applications.

\end{document}